\documentclass[journal]{IEEEtran}
\usepackage{cite}
\usepackage{url}
\usepackage{ragged2e}
\usepackage{epsfig}
\usepackage{graphicx}
\usepackage{mathptmx}      
\usepackage{amsmath,amssymb} 
\usepackage{subfigure}
\usepackage{algorithm}
\usepackage{algorithmicx}
\usepackage{algpseudocode}
\usepackage{graphics}
\usepackage{threeparttable}
\usepackage{color}
\usepackage[normalem]{ulem}
\usepackage{multirow}
\usepackage{float}
\usepackage{amsfonts}
\usepackage{bm}
\usepackage{array}
\usepackage[table]{xcolor}
\usepackage{colortbl}
\usepackage{pifont}
\usepackage{diagbox}
\usepackage{rotating}
\usepackage{booktabs}
\usepackage{overpic}
\usepackage{textcomp}
\usepackage{contour}
\usepackage{enumitem}

\usepackage[american]{babel}
\usepackage{microtype}
\usepackage{bbding}

\usepackage[pagebackref=false,breaklinks=true,colorlinks, bookmarks=false]{hyperref}

\usepackage{silence}
\def\ie{\emph{i.e.}}
\def\eg{\emph{e.g.}}
\def\etc{\emph{etc}}
\def\etal{{\em et al.~}}

\definecolor{bblue}{rgb}{0,150,230}
\definecolor{mygray}{gray}{.92}

\newcommand{\figref}[1]{Fig.~\ref{#1}}
\newcommand{\tabref}[1]{Table~\ref{#1}}
\newcommand{\eqnref}[1]{(Eq.~\ref{#1})}
\newcommand{\secref}[1]{$\S$ \ref{#1}}

\newcommand{\fdp}[1]{#1}

\newcommand{\AddText}[3]{\put(#1,#2){\contour{black}{\textbf{\textcolor{green}{#3}}}}}

\newcommand{\trb}[1]{\textbf{\textcolor{black}{#1}}}

\newcommand{\supp}[1]{\textcolor{magenta}{#1}}
\newcommand{\cmm}[1]{\textcolor{blue}{#1}}

\def\ourdataset{\textit{SIP}}
\graphicspath{{./Imgs/}{../../figure/}}
\begin{document}
%
\title{Rethinking RGB-D Salient Object Detection: Models, Data Sets, and Large-Scale Benchmarks}
%
%
%

\author{Deng-Ping~Fan,
        Zheng~Lin,
        Zhao~Zhang,
        Menglong Zhu,
        and~Ming-Ming~Cheng
\IEEEcompsocitemizethanks{
\IEEEcompsocthanksitem D.-P.~Fan, Z.~Lin, Z.~Zhang, and M.-M.~Cheng are with the College of Computer Science, Nankai University, Tianjin, China
\IEEEcompsocthanksitem M.~Zhu is with the Google AI, USA.
\IEEEcompsocthanksitem M.-M.~Cheng is the corresponding author (email: cmm@nankai.edu.cn).
}
\thanks{Manuscript received July 16, 2019; revised March 10, 2020.}}

\markboth{TRANSACTIONS ON NEURAL NETWORKS AND LEARNING SYSTEMS,~Vol.~X, No.~X, MONTH~2020}%
{Shell \MakeLowercase{\textit{et al.}}: Bare Demo of IEEEtran.cls for IEEE Journals}
%



\maketitle

\begin{abstract}
   \justifying
   The use of RGB-D information for salient object detection has been \fdp{extensively} explored in recent years.
   However, relatively few efforts have been \fdp{put towards}
   modelling salient object detection \fdp{in} real-world human activity scenes with RGB-D.
   In this work, we fill the gap by making the following contributions
   to RGB-D salient object detection.
   (1) We carefully collect a new SIP (salient person) dataset,
   which consists of $\sim$1K high-resolution images that cover diverse real-world scenes
   from various viewpoints, poses, occlusion\fdp{s}, illumination\fdp{s}, and background\fdp{s}.
   (2) \fdp{We conduct a large-scale (and, so far, the most comprehensive) benchmark}
   comparing contemporary methods,
   which has long been missing in the \fdp{field} and can serve as
   a baseline for future research.
   We systematically summarize 32 popular models, \fdp{and evaluate
   18 parts of 32 models on seven datasets containing a total of about 97K images.}
   (3) We propose a simple \fdp{general} architecture,
   called Deep Depth-Depurator Network (D$^3$Net).
   It consists of a depth depurator unit (\fdp{DDU}) and a three-stream feature learning module (\fdp{FLM}),
   \fdp{which performs} low-quality depth map filtering and
   cross-modal feature learning respectively.
   These components form a nested structure and are elaborately designed
   to be learned jointly.
   D$^3$Net exceeds the performance of any prior contenders across \fdp{all} five metrics
   \fdp{under consideration}, thus \fdp{serving} as a strong model to advance research \fdp{in this field.}
   We also demonstrate that D$^3$Net can be used to efficiently extract salient object
   masks from real scenes, enabling effective background changing application
   with \fdp{a speed of} \fdp{65fps} on a single GPU.
   All the saliency maps, our new SIP dataset, \fdp{the D$^3$Net} model, and \fdp{the} evaluation tools are publicly available at
   \url{https://github.com/DengPingFan/D3NetBenchmark}.
\end{abstract}

\begin{IEEEkeywords}
\fdp{Benchmark, SIP Dataset, Salient Object Detection, Saliency, RGB-D}.
\end{IEEEkeywords}

%
\IEEEpeerreviewmaketitle

\section{Introduction}\label{sec:introduction}
\IEEEPARstart{H}{ow} to take \fdp{high-quality} photos has become one of the most
important competition points among mobile phone manufacturers.
Salient object detection (SOD) methods~\cite{BorjiCVM2019,wang2018detect,
fu2015object,zhang2019salient,borji2015salient,wang2019revisiting,fan2019shifting,
zeng2019Multi,wu2019Mutual,zhang2019CapSal,feng2019Attentive,hu2018recurrently,
wang2018salient,xu2019saliency,liu2018picanet,deng2018r3net,HouPami19Dss,tao2015manifold}
have been \fdp{incorporated into mobile phones} and been widely used \fdp{for creating}
perfect portraits by automatically adding
large aperture and other enhancement effects.
While existing SOD methods~\cite{li2017instance,Islam2017Salient,luo2017non,
chen2016disc,lee2016deep,zhao2015saliency,chen2018reverse,Zhuge2019deep,
su2018selectivity,jiang2018super,li2019deep,jia2019richer,huang2017300,
li2018contour,Kummerer_2018_ECCV,Chen_2017_ICCV,amirul2018revisiting} have
achieved remarkable success, most of them only rely on RGB images and
ignore the important depth information, which is widely available in modern
\fdp{smartphones} (\eg, iPhone X, Huawei Mate10, and Samsung Galaxy S10).
\fdp{Thus,} fully utilizing RGB-D information for SOD detection has recently attracted
\fdp{significant} research attention~\cite{ciptadi2013depth,ju2014depth,cheng2014depth,
peng2014rgbd,ren2015exploiting,feng2016local,guo2016salient,han2017cnns,
qu2017rgbd,song2017depth,zhu2017innovative,liang2018stereoscopic,
zhu2018pdnet,chen2018progressively,wang2017stereoscopic,zhao2019Contrast}.

\begin{figure}[t!]
  \centering
  \small
  \begin{overpic}[width=\columnwidth]{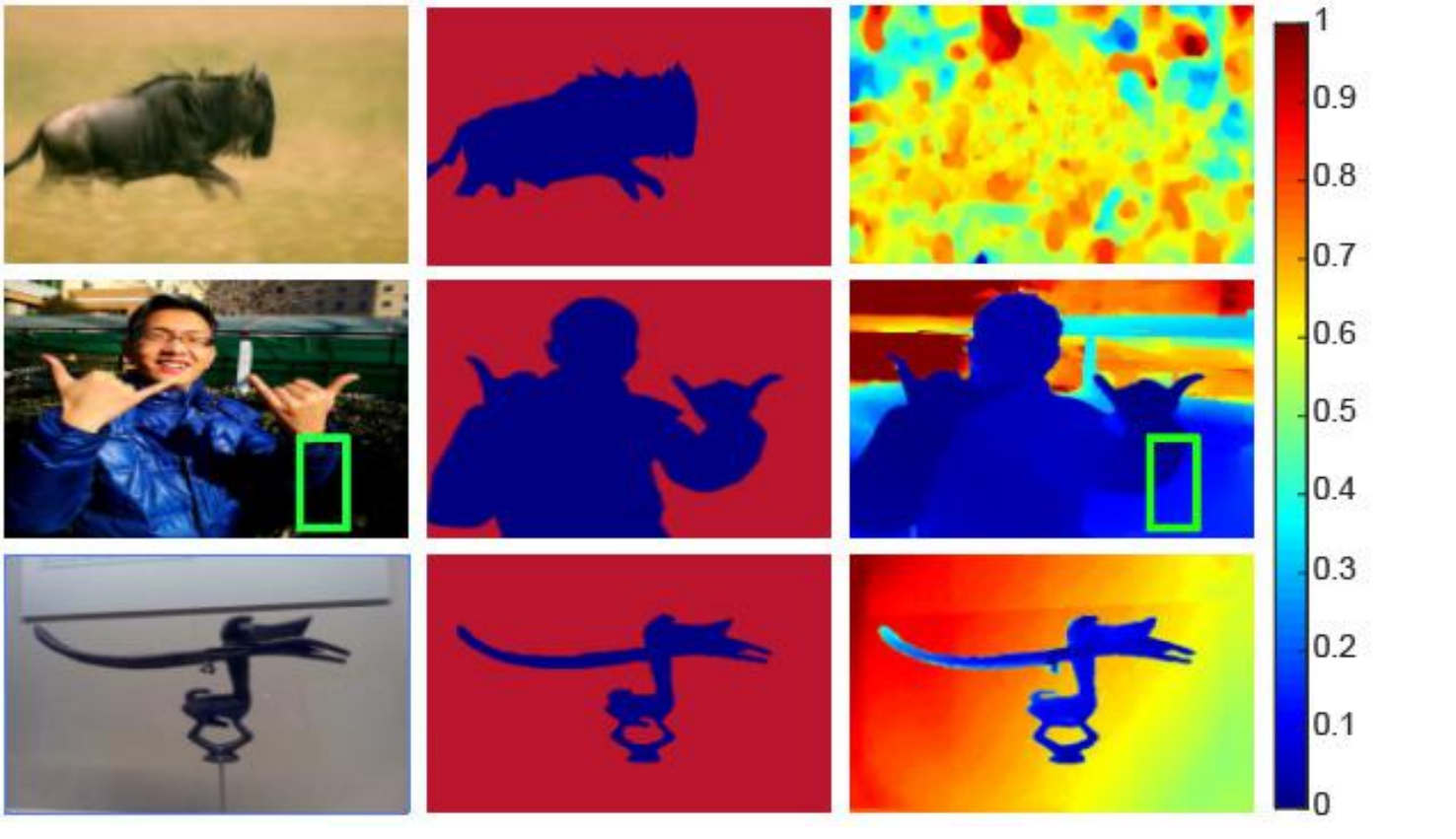}
  \AddText{65}{43}{\small {low}}
  \AddText{65}{23}{\small {mid}}
  \AddText{65}{3}{\small {high}}
  \end{overpic}
  \caption{\small
  Left to right: input image, ground truth, and the corresponding depth map.
  The quality of the depth map from low ($1^{st}$ row), mid ($2^{nd}$ row)
  to high (last row).
  As shown in the 2$^{nd}$ row, it is difficult to recognize the boundary of
  the human's arm in the boundary box region.
  However, it is clearly visible in the depth map.
  The high-quality depth maps benefit the RGB-D based
  salient object detection task.
  These three examples are from \fdp{the} NJU2K~\cite{ju2014depth}, our \textit{SIP} and NLPR~\cite{peng2014rgbd} datasets respectively.
  }\label{fig:DepthQuality}
\end{figure}

\begin{figure*}[t!]
  \centering
  \small
  \begin{overpic}[width=\textwidth]{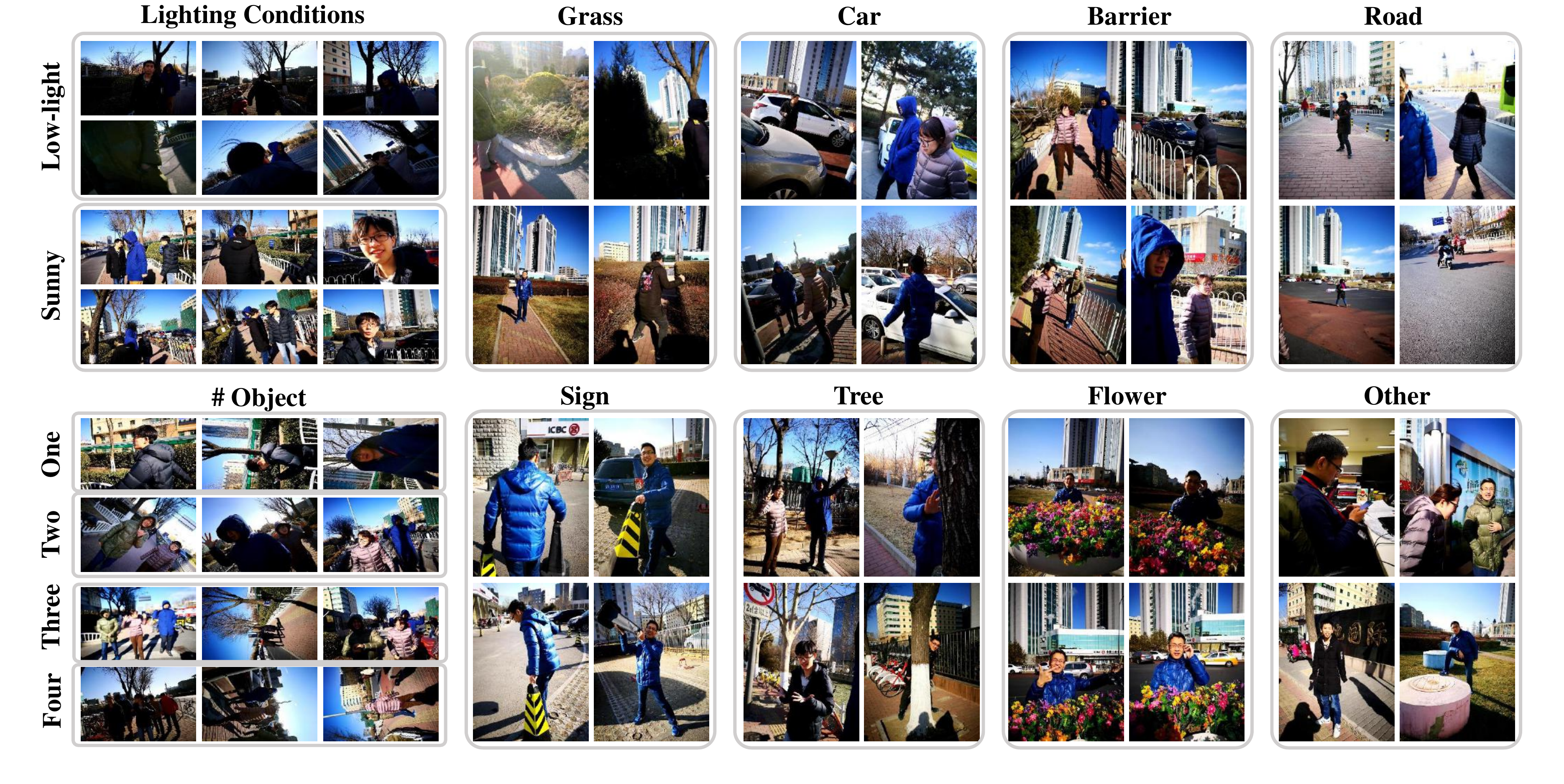}
  \end{overpic}
  \vspace{-20pt}
  \caption{
  \fdp{Representative subsets in our \ourdataset. The images in \ourdataset~are grouped into eight subsets according to background objects (\ie,
  grass, car, barrier, road, sign, tree, flower, and other), different lighting conditions (\ie, low-light, sunny with clear object boundary) and various number of objects (\ie, 1, 2, $\geq$3).}
  }\label{fig:RepresentativeScenarios}
\end{figure*}

\begin{figure*}[t!]
  \centering
  \small
  \begin{overpic}[width=\textwidth]{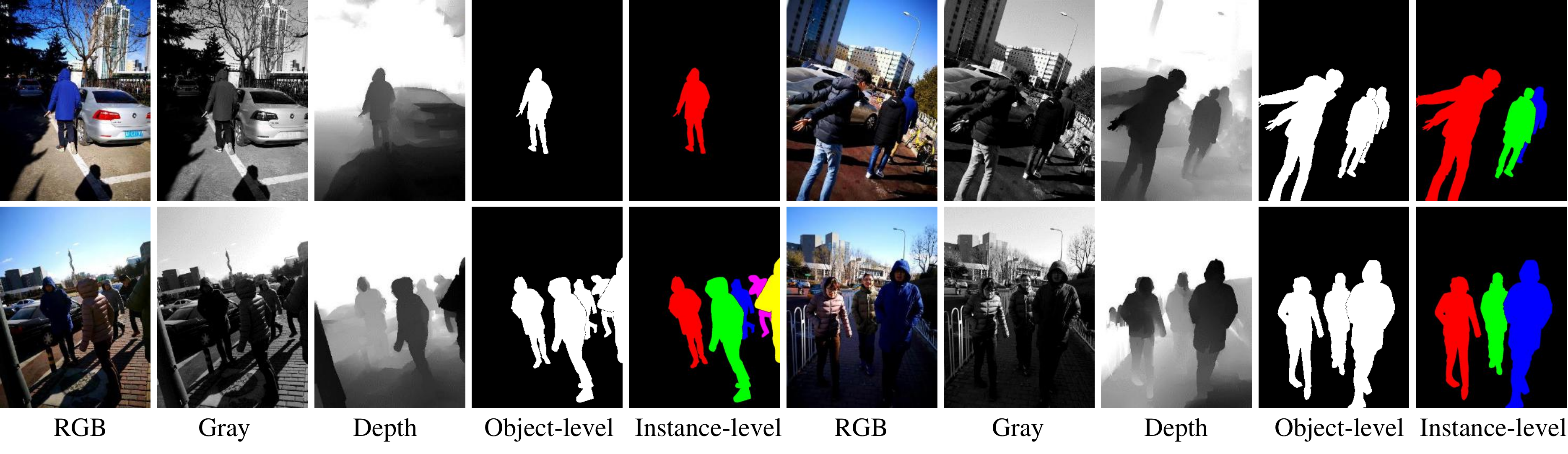}
  \end{overpic}
  \vspace{-10pt}
  \caption{Examples of images, depth maps and annotations (\ie, object-level, instance-level)
  in our \emph{SIP} dataset with different number\fdp{s} of salient objects, object size\fdp{s},
  object position\fdp{s}, scene \fdp{complexities}, and lighting conditions.
    Note that the ``RGB'' and ``Gray'' images are captured by two different
    monocular cameras \fdp{from short distances.}
    Thus, the ``Gray'' image\fdp{s} are slightly different from the grayscale image\fdp{s}
    obtained from colorful (RGB) image.
    Our \emph{SIP} dataset provides a new direction such as depth
    estimating from ``RGB'' and ``Gray'' image\fdp{s}, \fdp{and instance-level RGB-D salient object detection.}
  }\label{fig:SIPExamples}
  \vspace{20pt}
\end{figure*}

One of the primary goals of existing \fdp{smartphone cameras} is \fdp{to identify} humans
in visual scenes, \fdp{through either coarse, bounding-box-level, or instance-level; segmentation.}
To this end, intelligence solutions, such as RGB-D saliency detecting techniques have gained
considerable attention.


However, \fdp{most} existing RGB-D based SOD methods are tested on RGB-D images
taken by Kinect~\cite{zhang2012microsoft} or a light field camera~\cite{ng2005light},
or estimated by optical flow~\cite{liu2011sift}, which have different
characteristics from \emph{actual} smartphone cameras.
Since human\fdp{s are} the key subject\fdp{s} of \fdp{photographs taken with smartphones,}
a human-oriented RGB-D dataset \fdp{featuring} realistic, \fdp{in-the-wild images would} be
more \fdp{useful} for mobile manufacturers.
Despite the effort of some authors~\cite{ju2014depth,peng2014rgbd} to
augment their scenes with additional objects, a human-centered RGB-D dataset
for salient object detection does not yet exist.

Furthermore, although depth map\fdp{s provide} important complementary information
for identifying salient objects, \fdp{the low-quality versions often}
cause wrong detections~\cite{cong2016saliency}.
While existing RGB-D based SOD models typically fuse RGB and depth features by
different strategies~\cite{zhao2019Contrast}.
\fdp{There is no model that explicitly/automatically discard
the low-quality depth map in the RGB-D SOD field.}
%
We believe such models \fdp{have a high potential for driving} this field forward.


\fdp{In addition to} the limitations of current RGB-D datasets and models \fdp{already mentioned},
most RGB-D \fdp{studies} also suffer from \fdp{several other} common \fdp{constraints}, including:

\textbf{Sufficiency}. Only \fdp{a} limited \fdp{number of } datasets (1$\sim$4) \fdp{have been}
benchmarked in recent papers~\cite{peng2014rgbd,cong2018review} (\tabref{tab:ModelSummary}).
The \fdp{generalizability} of models cannot be properly accessed with such a small number of datasets.
%

\textbf{Completeness}. F-measure~\cite{achanta2009frequency}, MAE, and
PR \fdp{(precision \& recall) Curve} are the three \fdp{most} widely-used metrics in existing works.
However, as suggested by~\cite{fan2017structure,Fan2018Enhanced}, these metrics
essentially \fdp{act at a pixel-level}.
It is \fdp{thus} difficult to draw thorough and reliable conclusions
from quantitative evaluations~\cite{margolin2014evaluate}.

\textbf{Fairness}.
Some works~\cite{zhao2019Contrast,wang2019adaptive,chen2018progressively} use the
\emph{same} F-measure metric, \fdp{but} do not explicitly describe which
statistic (\eg, mean or max)
\fdp{was used}, easily resulting in unfair
comparison and inconsistent performance.
Meanwhile, the different threshold strategies \fdp{for} F-measure (\eg, 255 varied thresholds~\cite{wang2019adaptive,zhao2019Contrast,du2016improving},
adaptive saliency threshold~\cite{peng2014rgbd,feng2016local}, and self-adaptive threshold~\cite{han2017cnns})
will result in different performance.
\fdp{It is thus of crucial need to provide a fair comparison of} RGB-D based SOD models by extensively
evaluating them with same metrics on a standard leaderboard.

\begin{table*}[t!]
  \centering
  \scriptsize
  \renewcommand{\arraystretch}{1.0}
  \renewcommand{\tabcolsep}{1.0mm}
  \caption{Comparison of current RGB-D datasets in terms of year
  	(\textbf{Year}), publication (\textbf{Pub.}),
  	dataset size (\textbf{DS.}), number of objects in the images (\textbf{\#Obj.}),
	type of scene (\textbf{Types.}), depth sensor (\textbf{Sensor.}),
	depth quality (\textbf{DQ.},
	\eg, high-quality depth map suffers from less random noise.
  	See last row in \figref{fig:DepthQuality}), annotation quality (\textbf{AQ.},
  	see \figref{fig:labelQuality}),
    whether or not provide grayscale image from monocular camera (\textbf{GI.}),
    center bias (\textbf{CB.}, see \figref{fig:datasetStatistics} (a)-(b)),
    and resolution (in pixel).
    H \& W denote the height and width of the image, respectively.
  }\label{tab:DatasetSummary}
  \begin{tabular}{c|r||l|l|r|c|c|r|c|c|c|c|c}
    \hline\toprule
  No.& Dataset & Year & Pub. & DS. &\#Obj.&Types. & Sensor. & DQ. & AQ. &GI.&CB. &Resolution (H$\times$W)\\
    \midrule
    \midrule
  1 &\textit{STERE}~\cite{niu2012leveraging} & 2012&CVPR  & 1K     & $\sim$one & internet               & Stereo camera+sift flow~\cite{liu2011sift} &      & High & No & High & [251$\sim$1200]$\times$[222$\sim$900]\\
  2 &\textit{GIT}~\cite{ciptadi2013depth}    & 2013&BMVC  & 0.08K  & multiple  & home environment       & Microsoft Kinect~\cite{zhang2012microsoft} &      & High & No & Low  &640 $\times$ 480\\
  3 &\textit{LFSD}~\cite{li2014saliency}     & 2014&CVPR  & 0.1K   & one       & 60 indoor/40 outdoor   & Lytro Illum camera~\cite{ng2005light}      &      & High & No & High &360 $\times$ 360 \\
  4 &\textit{DES}~\cite{cheng2014depth}      & 2014&ICIMCS& 0.135K & one       & 135 indoor             & Microsoft Kinect~\cite{zhang2012microsoft} & High &      & No & High &640 $\times$ 480 \\
  5 &\textit{NLPR}~\cite{peng2014rgbd}       & 2014&ECCV  & 1K     & multiple  & indoor/outdoor         & Microsoft Kinect~\cite{zhang2012microsoft} & High &      & No & High &640 $\times$ 480, 480 $\times$ 640  \\
  6 &\textit{NJU2K}~\cite{ju2014depth}       & 2014&ICIP  & 1.985K & $\sim$one & 3D movie/internet/photo& FujiW3 camera+optical flow~\cite{sun2010secrets}& & High & No & High & [231$\sim$1213]$\times$[274$\sim$828]\\
  7 &\textit{SSD}~\cite{zhu2017three}        & 2017&ICCVW & 0.08K  & multiple  & three stereo movies    & Sun's optical flow~\cite{sun2010secrets}   &      &      & No & Low  & 960 $\times$ 1080 \\
    \midrule
  \rowcolor{mygray}
  8 &\textbf{\emph{SIP (Ours)}}              & \textbf{2020}&  TNNLS    & \textbf{0.929K}     & \textbf{multiple}  & \textbf{person in the wild}     & \textbf{Huawei Mate10}                              & \textbf{High} & \textbf{High} & \textbf{Yes} & \textbf{Low}  &\textbf{992$\times$744}\\
    \bottomrule
    \hline
  \end{tabular}
\end{table*}

\subsection{Contribution}
To \fdp{address} the \fdp{above-mentioned} problem\fdp{s},
we \fdp{provide} three distinct contributions.

(1) We \fdp{have} built a new \textbf{S}al\textbf{i}ent \textbf{P}erson
(\emph{\textbf{SIP}}) dataset (see \figref{fig:RepresentativeScenarios}, \figref{fig:SIPExamples}).
It consists of 929 accurately annotated high-resolution images which
are designed to contain multiple salient persons per image.
It \fdp{is} worth mentioning that the depth maps are captured by \fdp{a}
real \fdp{smartphone}. We believe such a dataset is highly \fdp{valuable}
and \fdp{will facilitate the application of} RGB-D model\fdp{s} to \fdp{mobile devices.}
%
Besides, the dataset is carefully \fdp{designed} to cover diverse scenes,
various challenging situations (\eg, occlusion, appearance change),
and elaborately annotated with pixel-level ground truth\fdp{s} (GT).
Another discriminative feature of our \emph{SIP} dataset is the
availability of both RGB and \fdp{grayscale} images captured by \fdp{a}
binocular camera, \fdp{which can benefit a broad number of} research directions,
\fdp{such as}, stereo matching, depth \fdp{estimation,} human-centered detection, \etc.

(2) With the proposed \emph{SIP} and \fdp{six existing} RGB-D
datasets~\cite{ju2014depth,
niu2012leveraging,cheng2014depth,peng2014rgbd,zhu2017three,li2014saliency},
we provide a more comprehensive \fdp{comparison} of 32 classical RGB-D salient
object detection models and present the \fdp{large-scale} ($\sim$97K images)
fair evaluation of 18 state-of-the-art (\fdp{SOTA}) algorithms~\cite{peng2014rgbd,cheng2014depth,ju2014depth,
ren2015exploiting,feng2016local,cong2016saliency,guo2016salient,zhu2017innovative,
qu2017rgbd,song2017depth,cong2017iterative,cong2018hscs,cong2018co,liang2018stereoscopic,
chen2018progressively,han2017cnns}, making \fdp{our study} a good all-around RGB-D benchmark.
To further promote the development of this field, we additionally provide an
online evaluation platform with the preserved test set.

(3) \fdp{We} propose a simple general model called
Deep Depth-Depurator Network (\textbf{D$^3$Net}), which
learns to automatically discard low-quality
depth map\fdp{s} using a novel depth depurator unit (DDU).
Thanks to the gate connection mechanism, our D$^3$Net
can predict salient object\fdp{s} accurately.
Extensive experiments \fdp{demonstrate} that our D$^3$Net
remarkably outperforms prior work on many \fdp{challenging} datasets.
Such a general framework design helps to learn cross-modality
features from RGB image\fdp{s} and depth map\fdp{s}.

Our contributions offer a systematic benchmark \fdp{equipped} with the basic tools for
comprehensive assessment of RGB-D models, offering deep insight into the task of RGB-D based modelling
and encouraging future research \fdp{in} this direction.

\subsection{Organization}
In \secref{sec:relatedWorks}, we first review \fdp{current} datasets for
RGB-D salient object detection, \fdp{as well as} representative \fdp{models for this task.}
Then, we present details \fdp{on the} proposed salient person dataset \ourdataset~in~\secref{sec:SIPdataset}.
In~\secref{sec:D3Net}, we describe our D$^3$Net model for RGB-D
salient object detection by \fdp{explicitly filtering out the low-quality depth maps.}


In \secref{sec:experiments}, we provide both \fdp{a} quantitative and qualitative
experimental analysis of the proposed algorithm. Specifically,
in \secref{sec:setting}, we offer more details on our experimental
settings, including \fdp{the} benchmarked models, datasets and runtime.
In \secref{sec:metrics}, five evaluation metrics (E-measure~\cite{Fan2018Enhanced}, S-measure~\cite{fan2017structure}, MAE, PR Curve, and F-measure~\cite{achanta2009frequency}) are described in detail.
In \secref{sec:metricStatistics}, \fdp{we provide the mean statistics over different datasets} and summarize them in \tabref{tab:BenchmarkResults}.
\fdp{comparison results of 18 SOTA RGB-D based SOD models over seven datasets}, namely \textit{STERE}~\cite{niu2012leveraging}, \textit{LFSD}~\cite{li2014saliency},
\textit{DES}~\cite{cheng2014depth}, \textit{NLPR}~\cite{peng2014rgbd}, \textit{NJU2K}~\cite{ju2014depth},
\textit{SSD}~\cite{zhu2017three}, and \emph{SIP (Ours)} clearly demonstrate the robustness and efficiency of our D$^3$Net model.
%
Further, in \secref{sec:PerformanceComparision},
we provide \fdp{a} performance comparison between traditional and deep models.
We also \fdp{discuss the experimental results in more depth.}
In \secref{sec:SOTAComparison}, we provide visualizations of the
results and \fdp{present} saliency maps generated \fdp{for} various challenging scenes.
In \secref{sec:application}, we discuss some potential applications about human activities and provide an interesting \fdp{and} realistic
use scenario of D$^3$Net in a background changing application.
To better understand the contributions of
\fdp{DDU} in the proposed D$^3$Net, in \secref{sec:discussion}, we present the
upper and lower bound of the DDU.
%
All in all, the extensive experimental
results clearly demonstrate that our D$^3$Net model exceeds the performance
of any prior competitors across \fdp{five} different metrics.
In \secref{sec:limitation}, we discuss the limitation\fdp{s} of this work.
Finally, \secref{sec:conclusion} concludes the paper.

\section{Related Works}\label{sec:relatedWorks}
\subsection{RGB-D Datasets}
Over the past few years, several RGB-D datasets
have been constructed for SOD. Some statistics of these datasets are
shown in \tabref{tab:DatasetSummary}.
\fdp{Specifically}, \fdp{the} \emph{STERE}~\cite{niu2012leveraging} dataset \fdp{was}
the first \fdp{collection of} stereoscopic photo\fdp{s} in this field.
%
\emph{GIT}~\cite{ciptadi2013depth}, \emph{LFSD}~\cite{li2014saliency}
and \emph{DES}~\cite{li2014saliency} are three \fdp{small-sized} datasets.
\emph{GIT} and \emph{LFSD} \fdp{were} designed \fdp{with} specific purposes in mind, \eg, saliency-based
segmentation of generic objects, and saliency detection on the light field.
\emph{DES} has 135 indoor images captured by Microsoft Kinect~\cite{zhang2012microsoft}.
Although \fdp{these} datasets \fdp{have} advanced the field to various degrees, they are
severely restricted by \fdp{their} small scale or low resolution.
%
To overcome \fdp{these} barriers, Peng~\etal created \emph{NLPR}~\cite{peng2014rgbd},
a \fdp{large-scale} RGB-D dataset with \fdp{a} resolution of 640$\times$480.
Later, Ju~\etal built \emph{NJU2K}~\cite{ju2014depth}, \fdp{which} has become one of
the most popular RGB-D datasets.
\fdp{The} recent \emph{SSD}~\cite{zhu2017three} dataset partially remedied the resolution
restriction of \emph{NLPR} and \emph{NJU2K}. 
\fdp{However, it only contains 80 images. Despite the progress made by existing RGB-D datasets, they
still suffer from the common limitation of not capturing depth maps in the real smartphones, making them
unsuitable for reflecting real environmental conditions (\eg, lighting or distance to object).}
%

\begin{table*}[t!]
  \centering
  \scriptsize
  \renewcommand{\arraystretch}{1.0}
  \renewcommand{\tabcolsep}{1.2mm}
  \vspace{-10pt}
  \caption{Comparison of 31 classical RGB-D based SOD algorithms and the proposed baseline (D$^3$Net).
  \textbf{Train/Val Set. (\#)} = Training or Validation Set:
  \emph{NLR} = \emph{NLPR}~\cite{peng2014rgbd}. \emph{NJU} = \emph{NJU2K}~\cite{ju2014depth}.
  \emph{MK} = \emph{MSRA10K}~\cite{ChengPAMI15}.
  \emph{O} = \emph{MK} + \emph{DUTS}~\cite{wang2017learning}.
  \textbf{Basic:}
  4Priors = 4 priors, \eg, Region, Background, Depth, and Surface Orientation Prior.
  IPT = Initialization Parameters Transfer.
  LGBS Priors = Local Contrast, Global Contrast, Background, and Spatial Prior.
  RFR~\cite{sauer2011accurate} = Random Forest Regressor.
  MCFM = Multi-constraint Feature Matching.
  CLP = Cross Label Propagation.
  \textbf{Type:} T = Traditional. D = Deep learning.
  \textbf{SP.} = SuperPixel: \fdp{Whether or not use the superpixel algorithm.}
  \textbf{E-measure:} The range of scores over the \fdp{seven} datasets in \tabref{tab:BenchmarkResults}.
  Evaluation tools:\supp{\url{https://github.com/DengPingFan/E-measure}}.
  }\label{tab:ModelSummary}
  \scriptsize
  \begin{tabular}{c|r|l|l|l|c|l|c|c|c}
  \hline\toprule
  No. & Model & Year & Pub. & Train/Val Set. (\#) & Test (\#)  & Basic & Type & SP. & E-measure$\uparrow$~\cite{Fan2018Enhanced}\\
  \midrule
  \midrule
  1 & LS~\cite{ciptadi2013depth}   & 2013 & BMVC  & Without training dataset  & One & Markov Random Field                        & T & \checkmark &  Not Available                 \\
  2 & RC~\cite{desingh2013depth}   & 2013 & BMVC  &Without training dataset & One & Region Contrast, SVM~\cite{chang2011libsvm}& T &  &   Not available               \\
  3 & LHM~\cite{peng2014rgbd}      & 2014 & ECCV  & Without training dataset & One & Multi-Context Contrast                     & T & \checkmark & 0.653$\sim$0.771 \\
  4 & DESM~\cite{cheng2014depth}   & 2014 & ICIMCS& Without training dataset & One & Color/Depth Contrast, Spatial Bias Prior   & T &            & 0.770$\sim$0.868 \\
  5 & ACSD~\cite{ju2014depth}      & 2014 & ICIP  &Without training dataset& One & Difference of Gaussian                     & T & \checkmark & 0.780$\sim$0.850 \\
  6 & SRDS~\cite{fan2014salient}   & 2014 & DSP   &Without training dataset & One & Weighted Color Contrast                    & T &            &     Not available \\
  7 & GP~\cite{ren2015exploiting}  & 2015 & CVPRW & Without training dataset & Two & Markov Random Field, 4Priors               & T & \checkmark & 0.670$\sim$0.824 \\
  8 & PRC~\cite{du2016improving}   & 2016 & Access& Without training dataset& Two & Region Classification, RFR                 & T &            &  Not available \\
  9& LBE~\cite{feng2016local}      & 2016 & CVPR  & Without training dataset& Two & Angular Density Component                  & T & \checkmark & 0.736$\sim$0.890 \\
  10& DCMC~\cite{cong2016saliency} & 2016 & SPL   & Without training dataset& Two & Depth Confidence, Compactness, Graph       & T & \checkmark & 0.743$\sim$0.856 \\
  11& SE~\cite{guo2016salient}     & 2016 & ICME  & Without training dataset & Two & Cellular Automata                          & T & \checkmark & 0.771$\sim$0.856 \\
  12& MCLP~\cite{cong2017iterative}& 2017 & Cybernetic&Without training dataset& Two & Addition, Deletion and Iteration Scheme & T & \checkmark & Not available \\
  13& TPF~\cite{zhu2017three}      & 2017 & ICCVW &Without training dataset& Four& Cellular Automata, Optical Flow             & T & \checkmark &    Not available \\
  14& CDCP~\cite{zhu2017innovative}& 2017 & ICCVW &Without training dataset& Two & Center-dark Channel Prior                   & T & \checkmark & 0.700$\sim$0.820 \\
  15& DF~\cite{qu2017rgbd}         & 2017 & TIP     &\emph{NLR} (0.75K) + \emph{NJU} (1.0K) & Three & Laplacian Propagation, LGBS Priors                    & D & \checkmark &0.759$\sim$0.880\\
  16& BED~\cite{shigematsu2017learning}& 2017&ICCVW &\emph{NLR} (0.80K) + \emph{NJU} (1.6K) + MK (9K) & Two &Background Enclosure Distribution                        & D & \checkmark & Not available\\
  17& MDSF~\cite{song2017depth}    & 2017 & TIP     &\emph{NLR} (0.50K) + \emph{NJU} (0.5K) & Two & SVM~\cite{chang2011libsvm}, RFR, Ultrametric Contour Map& T &            & 0.779$\sim$0.885\\
  18& MFF~\cite{wang2017rgb}       & 2017 & SPL   & Without training dataset& One & Minimum Barrier Distance, 3D prior          & T &            & Not available \\
  19& Review~\cite{cong2018review} & 2018 & TCSVT & Without training dataset& Two &     Without model introduced      & T &           & Not available\\
  20& HSCS~\cite{cong2018hscs}     & 2018 & TMM   &Without training dataset& Two & Hierarchical Sparsity, Energy Function      & T & \checkmark &    Not available\\
  21& ICS~\cite{cong2018co}        & 2018 & TIP   & Without training dataset& One & MCFM, CLP                                   & T & \checkmark & Not available\\
  22& CDB~\cite{liang2018stereoscopic}&2018&NC    & Without training dataset& One & Background Prior                            & T & \checkmark & 0.698$\sim$0.830 \\
  23& SCDL~\cite{huang2018rgbd}    & 2018 & DSP     &\emph{NLR} (0.75K) + \emph{NJU} (1.0K) & Two   & Silhouette Feature, Spatial Coherence Loss                     & D & &  Not available \\
  24& PCF~\cite{chen2018progressively}&2018 &CVPR   &\emph{NLR} (0.70K) + \emph{NJU} (1.5K) & Three &Complementarity-Aware Fusion module~\cite{chen2018progressively}& D & & 0.827$\sim$0.925\\
  25& CTMF~\cite{han2017cnns}      & 2018 & Cybernetic&\emph{NLR} (0.65K) + \emph{NJU} (1.4K) & Four  & HHA~\cite{gupta2014learning}, IPT, Hidden Structure Transfer   & D & & 0.829$\sim$0.932\\
  26& ACCF~\cite{chen2018attention}& 2018 & IROS    &\emph{NLR} (0.65K) + \emph{NJU} (1.4K) & Three & Attention-Aware                                                & D & &  Not available\\
  27& PDNet~\cite{zhu2018pdnet}    & 2019 & ICME    &\emph{NLR} (0.50K) + \emph{NJU} (1.5K) + \emph{O} (21K) & Five & Depth-Enhanced Net~\cite{zhu2018pdnet}  & D &  & Not available    \\
  28& AFNet~\cite{wang2019adaptive}& 2019 & Access   &\emph{NLR} (0.70K) + \emph{NJU} (1.5K) & Three & Switch map, Edge-Aware loss                                    & D & & 0.807$\sim$0.887\\
  29& MMCI~\cite{chen2019multi}    & 2019 & PR      &\emph{NLR} (0.70K) + \emph{NJU} (1.5K) & Three & HHA~\cite{gupta2014learning}, Dilated Convolutional            & D & & 0.839$\sim$0.928\\
  30& TANet~\cite{chen2019three}   & 2019 & TIP     &\emph{NLR} (0.70K) + \emph{NJU} (1.5K) & Three & Attention-Aware Multi-Modal Fusion                             & D & & 0.847$\sim$0.941\\
  31& CPFP~\cite{zhao2019Contrast} & 2019 & CVPR    &\emph{NLR} (0.70K) + \emph{NJU} (1.5K) & Five  & Contrast Prior, Fluid Pyramid                                      & D & & 0.852$\sim$0.932\\
  \midrule
  \rowcolor{mygray}
  32& \textbf{D$^3$Net (Ours)}     & 2020 &         &\emph{NLR} (0.70K) + \emph{NJU} (1.5K) & Seven & Depth Depurator Unit                                           & D & & 0.862$\sim$0.953\\
  \bottomrule
  \hline
  \end{tabular}
\end{table*}

Compared to previous datasets, the proposed \emph{SIP} dataset has three fundamental differences:
\begin{itemize}
\item
It includes 929 images with many challenging situations~\cite{fan2018salient}
(\eg, dark background, occlusion, appearance change, and out-of-view) from various outdoor scenarios.

\item
The RGB, \fdp{grayscale} image\fdp{s}, and estimated depth map\fdp{s} are captured by \fdp{a smartphone}
with a dual-camera.
Due to the predominant application of SOD to human subjects on mobile phones,
we \fdp{also} focus on \fdp{this} and thus and thus, for the first time, emphasize
the salient person\fdp{s} in the real-world scenes.

\item
A detailed quantitative analysis is presented \fdp{for} the quality of the dataset (\eg, center bias, object size distribution, \etc.), which \fdp{was} not carefully investigated in previous RGB-D based studies.
\end{itemize}


\subsection{RGB-D Models}
Traditional models rely heavily on hand-crafted features
(\eg, contrast~\cite{desingh2013depth,peng2014rgbd,cheng2014depth,fan2014salient},
shape~\cite{ciptadi2013depth}).
By embedding the classical principles (\eg, spatial bias~\cite{cheng2014depth},
center-dark channel~\cite{zhu2017innovative},
3D~\cite{wang2017rgb}, background~\cite{liang2018stereoscopic,ren2015exploiting}),
difference of Gaussian~\cite{ju2014depth},
region classification~\cite{du2016improving},
SVM~\cite{desingh2013depth,song2017depth},
graph knowledge~\cite{cong2016saliency},
cellular automata~\cite{guo2016salient},
and Markov random field~\cite{fan2014salient,ren2015exploiting},
these models show that \fdp{specific hand-crafted} features can lead to decent performance.
\fdp{Several studies have} also explored \fdp{methods}
of integrating RGB and depth feature\fdp{s} via
various \fdp{combination strategies}, \fdp{using, for instance, angular densities}~\cite{feng2016local},
random forest regressor\fdp{s}~\cite{du2016improving,song2017depth}, and minimum barrier distance\fdp{s}~\cite{wang2017rgb}.
More details are shown in~\tabref{tab:ModelSummary}.

To overcome the limited expression ability of hand-crafted features,
recent works~\cite{shigematsu2017learning,qu2017rgbd,huang2018rgbd,
han2017cnns,zhu2018pdnet,chen2018progressively,chen2018attention,wang2019adaptive,
chen2019multi,chen2019three,zhao2019Contrast}
\fdp{have} proposed to introduce CNNs to infer salient object\fdp{s} from RGB-D data.
BED~\cite{shigematsu2017learning} and DF~\cite{qu2017rgbd} are two pioneering works for
\fdp{this, which} introduced deep learning technology \fdp{into} the RGB-D based SOD task.
More recently, Huang~\etal developed a more efficient end-to-end model~\cite{huang2018rgbd} with a modified loss function.
To address the shortage of training data, Zhu~\etal~\cite{zhu2018pdnet}
presented a robust prior model with \fdp{a} guided \fdp{depth-enhancement} module for SOD.
\fdp{In addition}, Chen~\etal developed a series of novel approaches \fdp{for this field},
such as hidden structure transfer~\cite{han2017cnns}, \fdp{a} complementarity fusion module
~\cite{chen2018progressively}, \fdp{an} attention-aware component~\cite{chen2018attention,
chen2019three}, and dilated convolutions~\cite{chen2019multi}.
Nevertheless, these works, to \fdp{the best of our} knowledge, are dedicated to extracting
\fdp{general} depth feature\fdp{s}/information.

We argue that not all information in \fdp{a} depth map is informative for
SOD, and low-quality depth map\fdp{s} often \fdp{introduce significant noise}
($1^{st}$ row in \figref{fig:DepthQuality}).
\fdp{Thus}, we \fdp{instead} design a simple general framework D$^3$Net,
which is equipped with a depth-depurator unit to explicitly exclude low-quality depth
maps when learning complementary feature. 

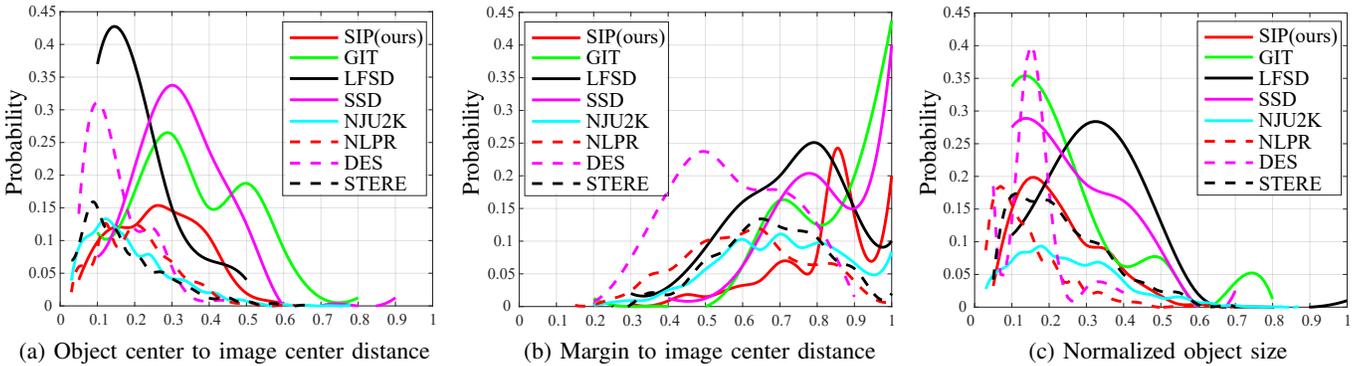
\begin{figure*}[t!]
  \small
  \centering
  \begin{overpic}[width=\textwidth]{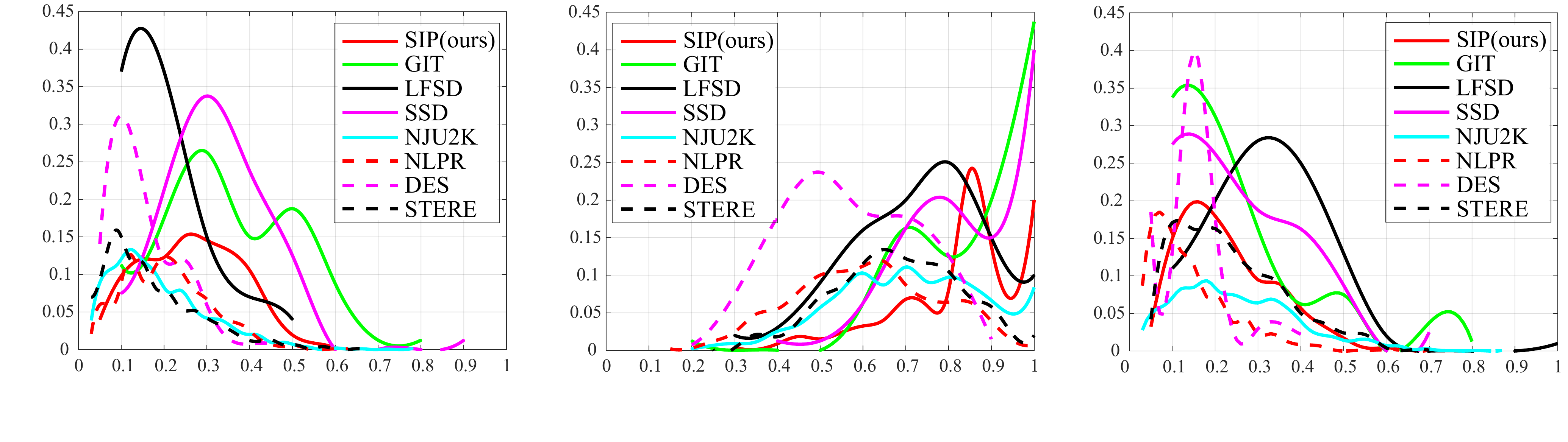}
  \put(2,2){(a) Object center to image center distance}
  \put(1,14){\rotatebox{90}{Probability}}
  \put(39,2){(b) Margin to image center distance}
  \put(34.5,14){\rotatebox{90}{Probability}}
  \put(76,2){(c) Normalized object size}
  \put(68,14){\rotatebox{90}{Probability}}
  \end{overpic}
  \caption{
  \small
  (a) Distribution of normalized object center distance from image center.
  (b) Distribution of normalized object margin (farthest point in an object)
  distance from image center.
  (c) Distribution of normalized object size.
  }\label{fig:datasetStatistics}
\end{figure*}

\begin{table*}[t!]
  \centering
  \small
  \renewcommand{\arraystretch}{1.0}
  \renewcommand{\tabcolsep}{2.6mm}
  \caption{\small Statistics regarding camera/object motions and salient object instance numbers in \emph{SIP}~dataset.}\label{tab:Pattern}
  \begin{tabular}{l||cccccccc||cc||ccccc}
  \hline\toprule
    & \multicolumn{8}{c||}{Background Objects}&\multicolumn{2}{c||}{Object Boundary}&\multicolumn{3}{c}{\# Object}\\
    \multirow{-2}{*}{\textbf{\ourdataset} (Ours)}& car & flower & grass & road & tree & signs  & barrier & other & dark & clear & 1 & 2 & $\geq$3\\
  \hline
    \#Img  & 107 & 9 & 154 & 140 & 97 & 25 & 366 & 32 & 162 & 767 & 591 & 159 & 179 \\
  \bottomrule
  \hline
  \end{tabular}
\end{table*}

\section{Proposed Dataset}\label{sec:SIPdataset}
\subsection{Dataset Overview}
We introduce \emph{SIP}, the first human activities oriented salient
person detection dataset. Our dataset contains 929 RGB-D images belonging to
\fdp{eight} different background scenes, \fdp{under two} different objecy boundary conditions,
\fdp{which portray} multiple actors. Each of them wears different clothes in
different images.
Following~\cite{fan2018salient}, the images are carefully selected to cover diverse
challenging cases (\eg, appearance change, occlusion, and shape complexity).
Examples can be found in \figref{fig:RepresentativeScenarios} and \figref{fig:SIPExamples}. The overall dataset
can be downloaded from our website \supp{\url{http://dpfan.net/SIPDataset/}}.

\subsection{Sensors and Data Acquisition}

\textbf{Image Collection:} We \fdp{used a} Huawei Mate 10 to collect our images.
The Mate 10's rear cameras feature high-grade Leica SUMMILUX-H lenses with bright
f/1.6 apertures
and combine 12MP RGB and 20MP Monochrome (grayscale) sensors.
The depth map is automatically estimated by the Mate10.
We asked nine people, all dressed in different colors,
to perform \fdp{specific} actions in real-world daily scenes.
Instructions on how to perform the action to cover different challenging situations
(\eg, occlusion, out-of-view) were given, \fdp{but} no instructions \fdp{on} style,
angle, or speed were provided, \fdp{in order to} record realistic data.

\textbf{Data Annotation:}
After capturing 5,269 images and the corresponding depth maps,
we first manually selected about 2,500 images, each of which included
one or multiple salient people. Following many famous SOD
datasets~\cite{achanta2009frequency,alpert2007image,ChengPAMI15,
JointSalExist17,li2017instance,li2015visual,LiuSZTS07Learn,
2001iccvSOD,wang2017learning,xia2017and,yan2013hierarchical},
%
six viewers \fdp{were} further instructed to draw the bounding boxes (bboxes)
\fdp{around} the most attention-grabbing person, \fdp{according to their first instinct.}
We adopt\fdp{ed} the voting scheme described in~\cite{peng2014rgbd} to discard images
with low voting consistency and chose top 1,000 \fdp{most} satisfactory images.
Another five annotators \fdp{were then} introduced to 
label accurate silhouettes of the salient objects according to the bboxes. We discard some images with low-quality annotations and finally obtained the 929 images with high-quality ground-truth annotations.

\subsection{Dataset Statistics}\label{sec:datasetStatistics}

\textbf{Center Bias:}
Center bias has been identified
as one of the most significant bias\fdp{es} of saliency detection datasets~\cite{li2014secrets}.
It \fdp{occurs because} subjects \fdp{tend to} look at the center of \fdp{a} screen~\cite{tatler2005visual}.
As \fdp{noted} in~\cite{fan2018salient}, simply overlap\fdp{ping} all of the maps in the dataset
\fdp{cannot well} describe the degree of center bias.

Following~\cite{fan2018salient}, we present the
statistics of two distance $R_o$ and $R_m$ in \figref{fig:datasetStatistics} (a \& b),
where $R_o$ and $R_m$ indicate how far an object center and margin (farthest)
point in an object are from the image center, respectively.
The center bias\fdp{es} of our \emph{SIP} and existing~\cite{niu2012leveraging,
ciptadi2013depth,li2014saliency,cheng2014depth,peng2014rgbd,ju2014depth,zhu2017three} datasets
are shown in \figref{fig:datasetStatistics} (a \& b).
Except for our \emph{SIP} and two small-scale datasets (\emph{GIT} and \emph{SSD}),
most datasets present a high degree of center bias,
\ie~the center of the object is close to the image center.

\begin{figure*}[t!]
  \centering
  \begin{overpic}[width=\textwidth]{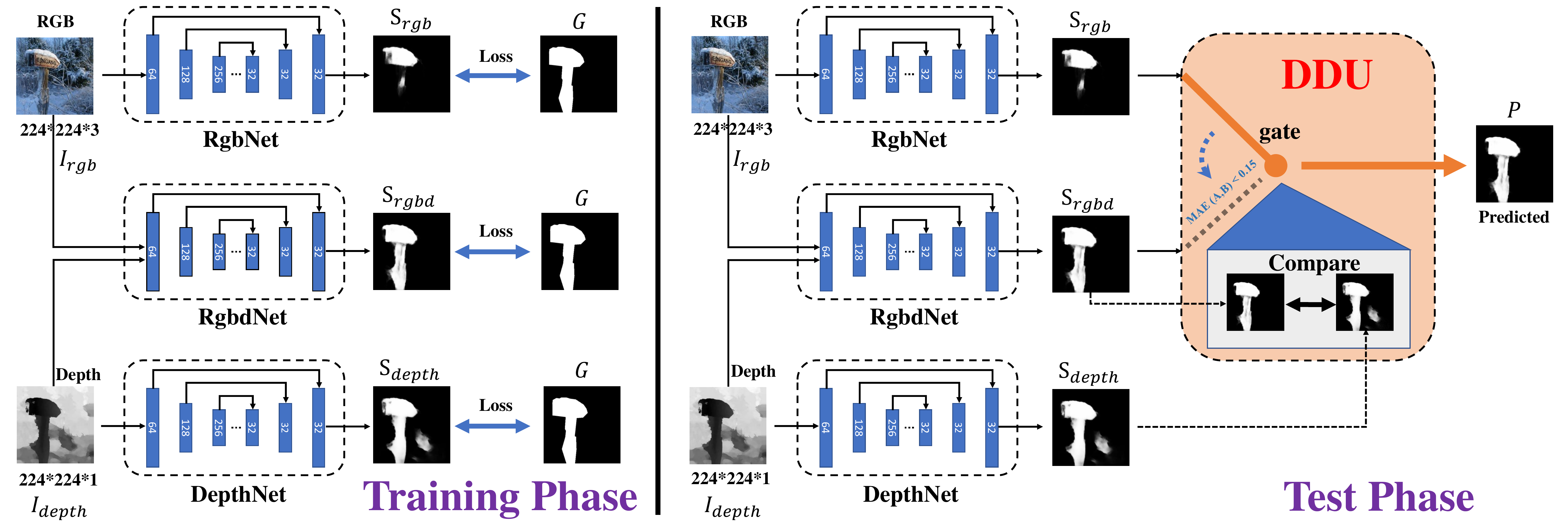}
  \end{overpic}
  \caption{\small \fdp{Illustration of the proposed D$^3$Net.
  In the training stage (Left), the input RGB and depth images are processed with three parallel sub-networks, \eg, RgbNet, RgbdNet, and DepthNet. The three sub-networks are based on a same modified structure of Feature Pyramid Networks (FPN) (see \secref{sec:FPN} for details).
  We introduced these sub-networks to obtain three saliency maps (\ie, $S_{rgb}$, $S_{rgbd}$, and $S_{depth}$) which considered both coarse and fine details of the input.
  In the test phase (Right), a novel depth depurator unit (DDU) (\secref{sec:DDU}) is utilized for the first time in this work to explicitly discard (\ie, $S_{rgbd}$) or keep (\ie, $S_{rgbd}$) the saliency map introduced by the depth map.}
  %
  %
   \fdp{In the training/test phase, these components form a nested structure and are elaborately designed (\eg, gate connection in DDU) to automatically learn the salient object from the RGB image and Depth image jointly.}
  }\label{fig:framework}
\end{figure*}

\textbf{Size of Object\fdp{s}:} We define object size as
the ratio of salient object pixels to \fdp{the total number of pixels in the image}.
\fdp{The distribution (\figref{fig:datasetStatistics} (c)) of normalized object size in \emph{SIP} are
0.48\%$\sim$66.85\% (avg.: 20.43\%).}

\textbf{Background Objects:} As summarized in \tabref{tab:Pattern},
\emph{SIP} includes diverse background objects (\eg, car\fdp{s}, tree\fdp{s}, and grass).
Models tested on such \fdp{a} dataset \fdp{would likely be able to} handle realistic scenes better
and thus \fdp{be} more practical.

\textbf{Object boundary Conditions:} In \tabref{tab:Pattern},
we show different object boundary conditions (\eg, dark and clear) in our \emph{SIP} dataset.
One example of \fdp{a} dark condition , \fdp{which often occurs in daily scenes}, can be found in \figref{fig:SIPExamples}.
\fdp{The depth maps obtained in low-light conditions inevitably introduce
more challenges for detecting salient objects.}

%
%

\textbf{\fdp{Number} of Salient Object:}
From \tabref{tab:DatasetSummary}, we note that existing datasets
fall short \fdp{in their} numbers of salient objects \fdp{(\eg, they often only have one)}.
Previous studies~\cite{kaufman1949discrimination}, however, \fdp{have shown}
that human\fdp{s} \fdp{can} accurately enumerate up to at least \fdp{five} objects without counting.
Thus, our \emph{SIP} is \fdp{designed to contain up to five salient objects per-image.}
The statistics of labelled objects in each image are shown in \tabref{tab:Pattern}
(\# Object).


\section{Proposed Model}\label{sec:D3Net}
According to motivation described in \figref{fig:DepthQuality}, cross-modality feature extraction and depth filter unit are highly desired; therefore we  proposed the simple general D$^3$Net \fdp{model} (illustrated in \figref{fig:framework}) which contains two components, \eg, a \emph{three-stream feature learning module} (\secref{sec:FPN}) and a \emph{depth depurator unit} (\secref{sec:DDU}).
The FLM (feature learning module) is utilized to extract the features from different modality. While the DDU (depth depurator unit) is \fdp{acting as a gate to} explicitly \fdp{filter out} the low-quality depth \fdp{maps}. \fdp{If DDU decides to filter out this depth map, the data flow will pass along with the RgbNet.}
These components form a nested structure,
\fdp{and} are elaborately designed to achieve robust performance
and high generalization ability on various challenging datasets.

\subsection{Feature Learning Module}\label{sec:FPN}
\fdp{Most existing models~\cite{kirillov2019panoptic,chen2019tensormask,xiong2019upsnet} have shown significant improvement for object detectors in several applications.
These models typically share a common structure of Feature Pyramid Networks (FPN)~\cite{lin2017feature}. Based on this motivation, we decide to introduce this component like FPN in our D$^3$Net baseline to efficiently extract the features in a pyramid manner. The entire D$^3$Net model is divided into the training phase and test phase due to the DDU has opted to use only in test phase.}

\fdp{As shown in~\figref{fig:framework}, the designed FLM appears in training and test phases.
The FLM consists of three sub-networks, \ie,\textit{ RgbNet, RgbdNet}, and \textit{DepthNet}.
Note that the three sub-networks have the same structure while fed with different input channel.
\fdp{Specifically, each sub-network receives a re-scaled image $I \in \{I_{rgb}, I_{rgbd}, I_{depth}\}$ with 224$\times$224 resolution.} The goal of FLM is to obtain the corresponding
predicted map S $\in \{S_{rgb}, S_{rgbd}, S_{depth}\}$.}

\vspace{5pt}
\fdp{As in~\cite{lin2017feature}, we also use bottom-up, top-down pathway, and lateral connections to extract the features. Then the outputs will be proportionally organized at multiple levels.}
\fdp{The FPN is independent of the backbone, thus for simplicity, we
adopt the VGG-16~\cite{simonyan2015very} architecture as our basic convolutional network to extract spatial features,
while utilizing more powerful backbone \cite{pami20Res2net}
feature extractor could be explored in future.
Some studies like~\cite{zhao2019pyramid} have shown that deeper layers retain more semantic information for locating objects. Based on this observation, we introduce a layer containing two 3$\times$3 convolution kernels on the basis of the 5 layers VGG-16 structure to achieve this goal.}

\begin{figure}[t!]
  \centering
  \begin{overpic}[width=\columnwidth]{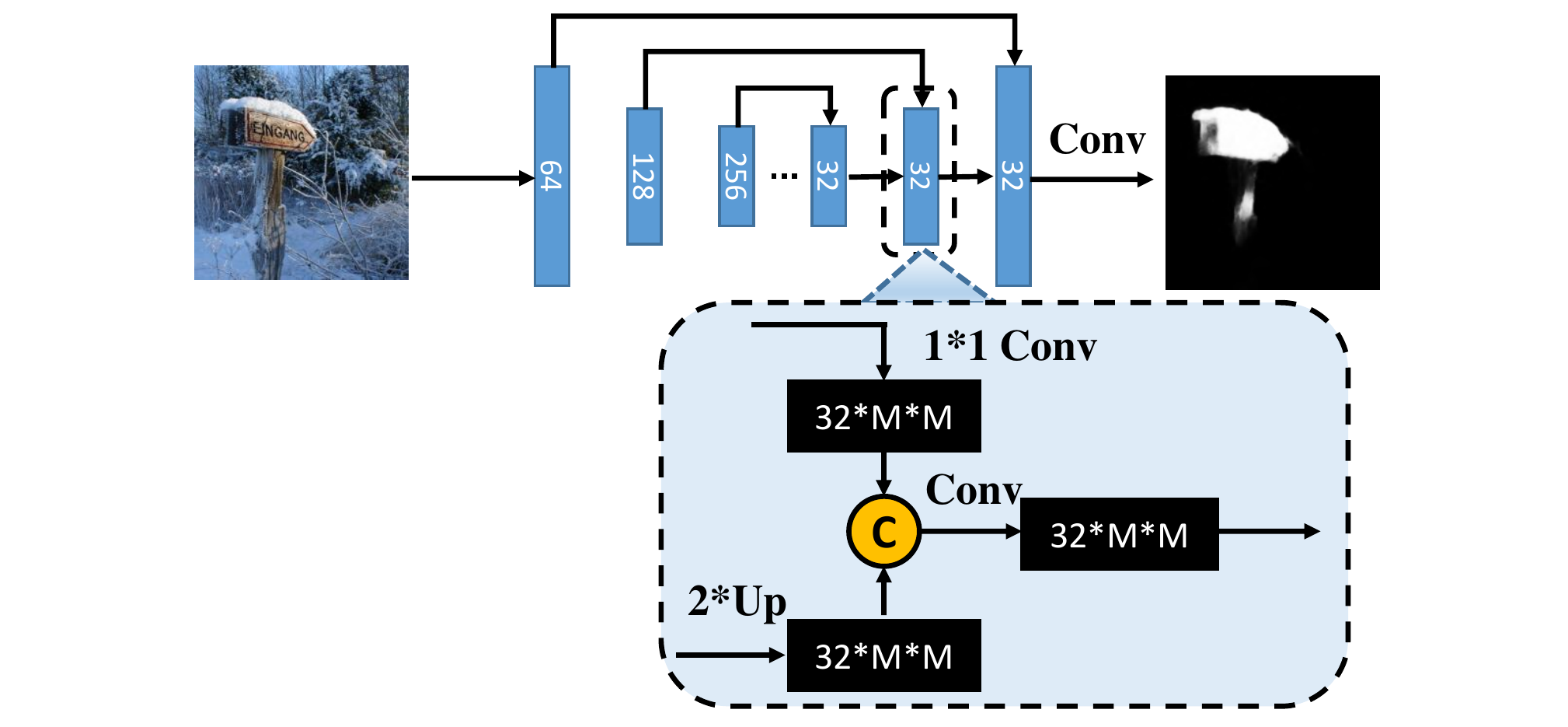}
  \end{overpic}
  \caption{\small The Feature Pyramid Network (FPN) is introduced to extract the context-aware information. Different from~\cite{lin2017feature}, we further add the sixth layer on the base of VGG-16 and the information merge strategy is concatenation rather than addition. More details can be found in \secref{sec:FPN}.
  }\label{fig:FPN}
\end{figure}

\fdp{As shown in \figref{fig:FPN}, our top-down features are built. For a specific layer (\eg, coarser layer), we first conduct a 2 $\times$ upsampling using nearest neighbor operation. Then, the upsampled feature is concatenated with the finer feature map to obtain rich features. Before concatenated with coarse map, the finer map undergoes a 1$\times$1 Conv operation to reduce the channel. For example, let $\mathbf{I_{rgbd}}\!\in\!\mathbb{R}^{W\!\times\!H\!\times\!4}$ denotes the four-dimensional feature tensor of the input of RgbdNet.
Then we define a set of anchors on different layers so that we can obtain a set of pyramid
feature tensors
with $C_i\times W_i\times H_i$, \ie, \{$64\times224\times224$, $128\times112\times112$, $256\times56\times56$, $512\times28\times28$, $512\times14\times14$, $32\times7\times7$, $32\times14\times14$, $32\times28\times28$, $32\times56\times56$, $32\times112\times112$, $32\times224\times224$\} on \{$F_i$, i$\in$ [1,11]\}, respectively. Note that the \{$F_1$, $F_2$, $F_3$, $F_4$, $F_5$\} are corresponding to the five convoluational stages of VGG-16 (\ie, \{$C_1$, $C_2$, $C_3$, $C_4$, $C_5$\}).}

\subsection{Depth Depurator Unit (DDU)}\label{sec:DDU}
In the test phase, we further adopt a new \textit{gate connection} strategy to obtain the optimal predicted map.  Low-quality depth map\fdp{s introduce} more noise than \fdp{informative cues} to the prediction. The goal of gate connection is to classify depth maps into reasonable and \fdp{low-quality} ones and not use the poor ones in the pipeline.

\begin{figure}[t!]
  \centering
  \small
  \begin{overpic}[width=\columnwidth]{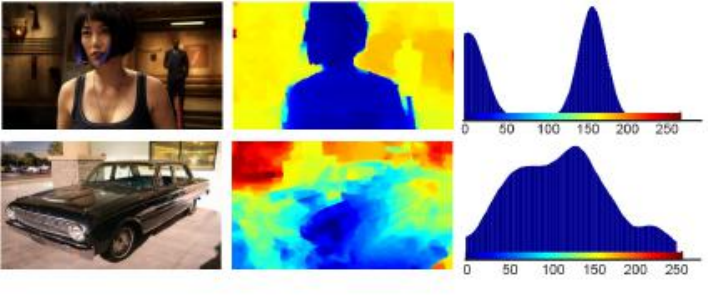}
   \put(7,0){(a) RGB}
   \put(40,0){(b) Depth}
   \put(67,0){(c) Histogram}
  \end{overpic}
  \caption{\small The smoothed histogram (c) of high-quality ($1^{st}$ row),
  	low-quality ($2^{nd}$ row) depth map, respectively.
  }\label{fig:DDUExample}
\end{figure}

As illustrated in \figref{fig:DDUExample} (b),
a stand-alone salient object in \fdp{a} high-quality depth map
is typically characterized by well-defined closed
boundaries and shows clear double peaks in its depth distribution.
%
The statistics of the depth maps in existing datasets~\cite{niu2012leveraging,
li2014saliency,cheng2014depth,peng2014rgbd,ju2014depth,zhu2017three}~
also support \fdp{the fact that} ``\emph{high quality depth maps usually
contain clear objects, \fdp{while} the elements in low-quality depth maps are
cluttered (2$^{nd}$ row in \figref{fig:DDUExample})}''.
In order to reject the low-quality depth maps, we propose DDU as \fdp{follows}:

\fdp{More specifically, in the test phase, the RGB and depth map \fdp{is} firstly re-sized to a fixed size (\eg, same as the training phase 224$\times$224) to reduce the computational complexity.
As shown in \figref{fig:framework} (right), the DDU is implemented with a gate connection. Denote the input images with three predicted maps $\textbf{S} \in \{S_{rgb}, S_{rgbd}, S_{depth}\}$, then the goal of DDU is to decide which predicted map $\textbf{P}\in [0,1] ^{W\times H}$ is optimal.
\begin{equation}
\textbf{P} =  F_{ddu} (\{S_{rgb}, S_{rgbd}, S_{depth}\}).
\label{eq:DDU}
\end{equation}}

Intuitively, there are two ways to achieve this goal, \eg, post-processing and pre-processing.
%
We propose a simple but general post-processing scheme for DDU.
The DDU is considered in the test phase rather than in the training phase.
Specially, a comparison unit $F_{cu}$ is leveraged to assess the similarity
between the $S_{depth}$ and $S_{rgbd}$ generated from DepthNet and RgbdNet,
respectively.
\begin{equation}
\begin{split}
F_{cu}=
\begin{cases}
1, & \delta(S_{rgbd}, S_{depth}) \leq t \\
0, & otherwise,
\end{cases}
\end{split}\label{eq:CU}
\end{equation}
where the $\delta(\cdot)$ represents distance function, and $t$ indicates a fixed threshold.
Note that the comparison unit $F_{cu}$ is act as an index to decide which sub-network
(RgbNet or RgbdNet) should be utilized.

\fdp{The key of our comparison unit is the DDU.
We utilize the comparison unit $F_{cu}$ as a gate connection to decide the
final/optimal predicted map $\textbf{P}$. Thus, our $F_{ddu}$ module can be formulated as:
\begin{equation}
\textbf{P} =  F_{cu}\cdot S_{rgbd} + \Bar{F}_{cu}\cdot S_{rgb},
\label{eq:DDU-CU}
\end{equation}
where $\Bar{F}_{cu} = 1 - F_{cu}$. The $F_{cu}$ can be viewed as a fixed weight. A more elegant formulation (adaptive weight) would be a part of our future work.}

\subsection{Implementation Details}
\fdp{\textbf{DDU.} The key component of our D$^3$Net is the DDU. In this work, we show a simple
yet powerful distance function formulated in \eqnref{eq:CU}. We leverage the mean absolute error (MAE) metric (same as \eqnref{eq:MAE}) to assess the distance between two maps. The basic motivation is that if the high-quality depth contains clear objects the DepthNet will easily detect these objects in $S_{depth}$ (see first row in \figref{fig:DDUExample}). The higher the quality of depth map in $I_{depth}$, the more similarity between the $S_{rgbd}$ and the $S_{depth}$. In other words, the predicted map $S_{rgbd}$ from RgbdNet have considered the feature from $I_{depth}$. If the quality of the depth map is low, then the predicted map from RgbdNet will quite different from the generated map from DepthNet. We have tested a set of values of the fixed threshold $t$ in \eqnref{eq:CU} such as, 0.01, 0.02, 0.05, 0.10, 0.15, 0.20, but have found $t = 0.15$ achieve the best performance.}

\textbf{Loss Function.}
We adopt the widely-used cross entropy loss function $L$ to train our model:
\begin{equation}
    \begin{aligned}
  L(\textbf{S}, \textbf{G}) = -\frac{1}{N} \sum\nolimits_{i=1}^N\Big(g_{i}\log(s_{i}) + (1 - g_{i})\log(1-s_{i})\Big),
    \end{aligned}
       \label{eq:4}
\end{equation}
where $\textbf{S}\in [0,1]^{224\times224}$ and $\textbf{G}\in \{0,1\}^{224\times224}$
\fdp{indicate} the estimated saliency map (\ie, $S_{rgb}$, $S_{rgbd}$, or $S_{depth}$) and the GT map, respectively. $g_i\in\textbf{G}$, $s_i\in\textbf{S}$, and $N$ denotes the total number of pixels.

\textbf{Training Settings.}
For fair comparisons, we follow the same training settings described in~\cite{zhao2019Contrast}.
We select 1485 image pairs from the \textit{NJU2K}~\cite{ju2014depth} and 700 image pairs from \textit{NLPR}~\cite{peng2014rgbd} dataset, respectively,  as the training data (Please refer to our website for the \textit{Trainlist.txt}).
The proposed D$^3$Net is implemented using Python, with the Pytorch toolbox.
We adopt Adam as the optimizer and the initial learning rate is 1e-4 and batchsize is set to 8. The total training is 30 epoch on
\fdp{a} GTX TITAN X GPU with 12G \fdp{of} memory.


\textbf{Data Augmentation.}
Due to the limited scale of existing datasets,
we augment the training samples by flipped the images horizontally
to overcome the risk of overfitting.

\section{Benchmarking Evaluation Results}\label{sec:experiments}
We benchmark about 97K \fdp{images} (5,398 images $\times$ 18 models)
in this study, making it the largest and most
comprehensive RGB-D based SOD benchmark to date.

\subsection{Experimental Settings}\label{sec:setting}
\textbf{Models.}
We benchmark 18 SOTA 
models (see \tabref{tab:BenchmarkResults}), including 10
traditional and 8 CNN based models.

\textbf{Datasets.}
We conduct our experiments on seven datasets (see \tabref{tab:BenchmarkResults}).
The test sets 
of \textit{NJU2K}~\cite{ju2014depth} and \textit{NLPR}~\cite{peng2014rgbd}
datasets, and the whole \textit{STERE}~\cite{niu2012leveraging}, \textit{DES}~\cite{cheng2014depth},
\textit{SSD}~\cite{zhu2017three}, \textit{LFSD}~\cite{li2014saliency},
and \textit{SIP} datasets are used for testing.

\textbf{Runtime.} In \tabref{tab:BenchmarkResults}, we summarize
the runtime of existing approaches.
The timings are tested on the same platform:
Intel Xeon(R) E5-2676v3 \@2.4GHz$\times$24 and GTX TITAN X.
Since~\cite{han2017cnns,chen2018attention,chen2019multi,chen2019three,
chen2018progressively,liang2018stereoscopic,cong2018hscs,cong2018co,cong2017iterative}
\fdp{have} not \fdp{released their codes}, the timings are borrowed from \fdp{the} original
paper\fdp{s} or provided by \fdp{the} authors.
Our D$^3$Net does not apply post-processing (\eg, CRF), thus the
computation only takes about 0.015s for a $224\times224$ image.

\begin{figure*}[t!]
  \centering
  \begin{overpic}[width=\textwidth]{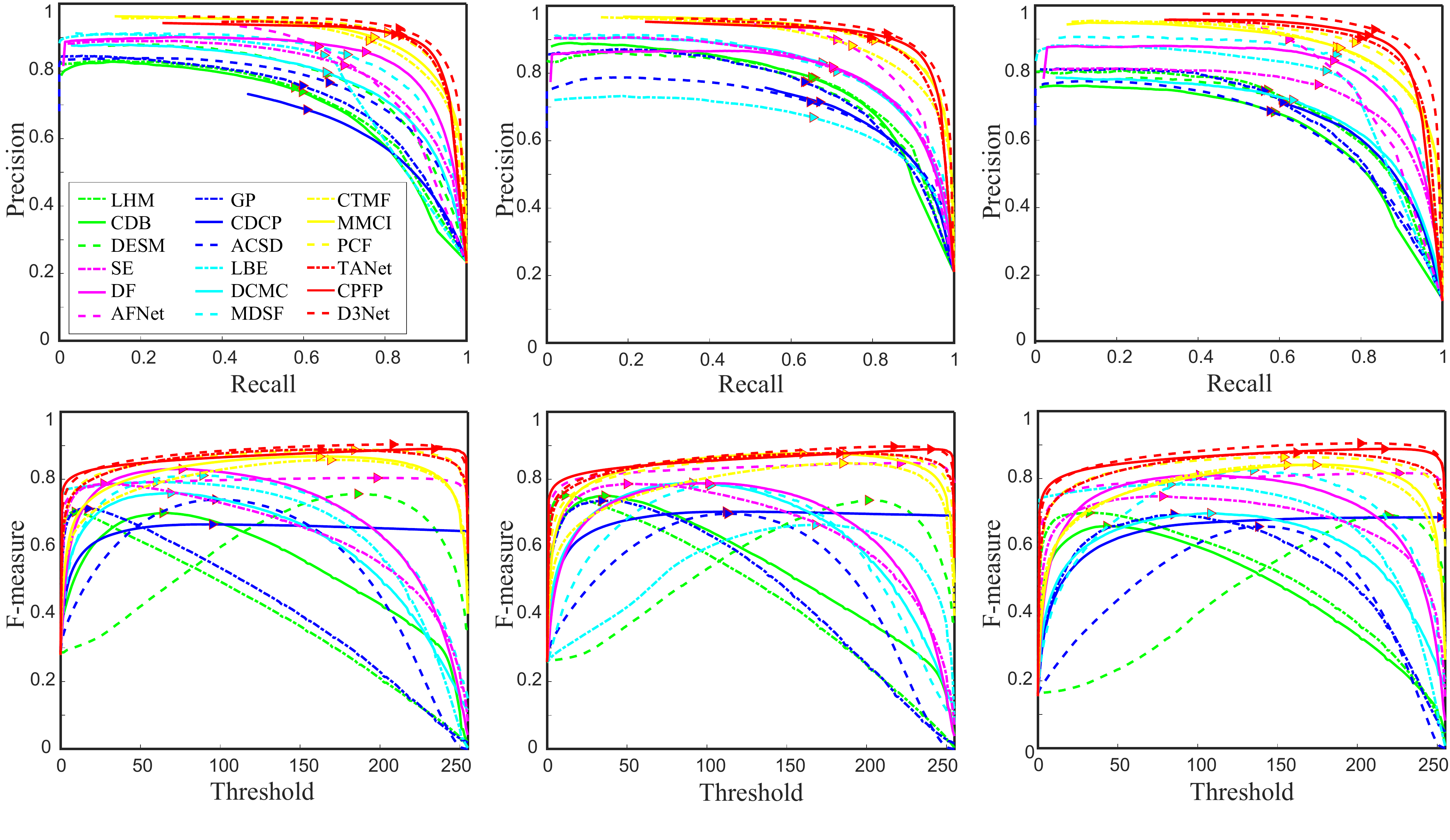}
    \put(12, 0){(a) NJU2K \cite{ju2014depth}}
    \put(46, 0){(b) STERE \cite{niu2012leveraging}}
    \put(80, 0){(c) NLPR \cite{peng2014rgbd}}
  \end{overpic}
  \caption{\small PR Curve (top) and F-measures (bottom) for
    \cmm{18 methods on NJU2K, STERE, and NLPR datasets},
    using \fdp{various} fixed thresholds.
  }\label{fig:PRCurveFmeasureOverall}
\end{figure*}



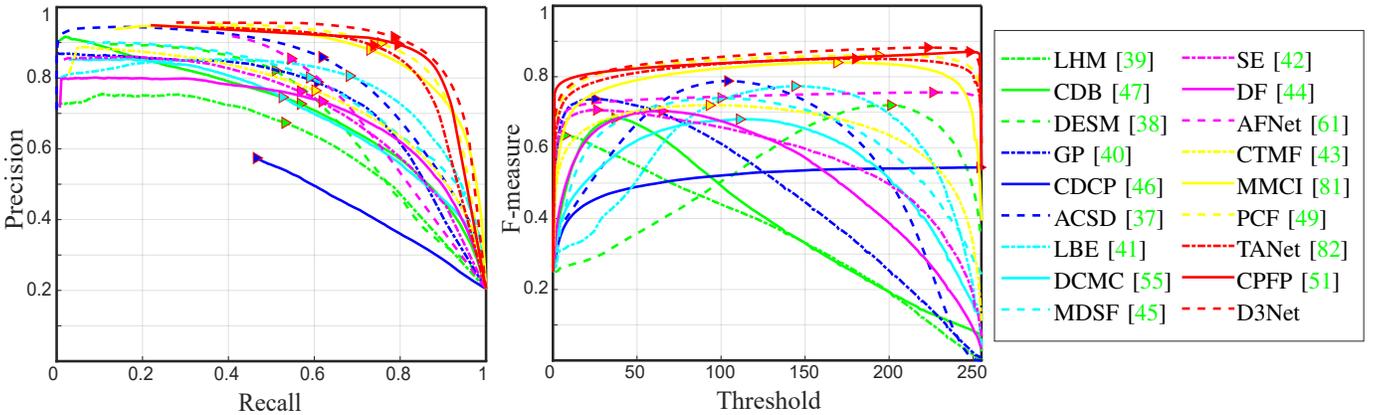
\begin{figure*}[t!]
  \centering
  \small
  \begin{overpic}[width=\linewidth]{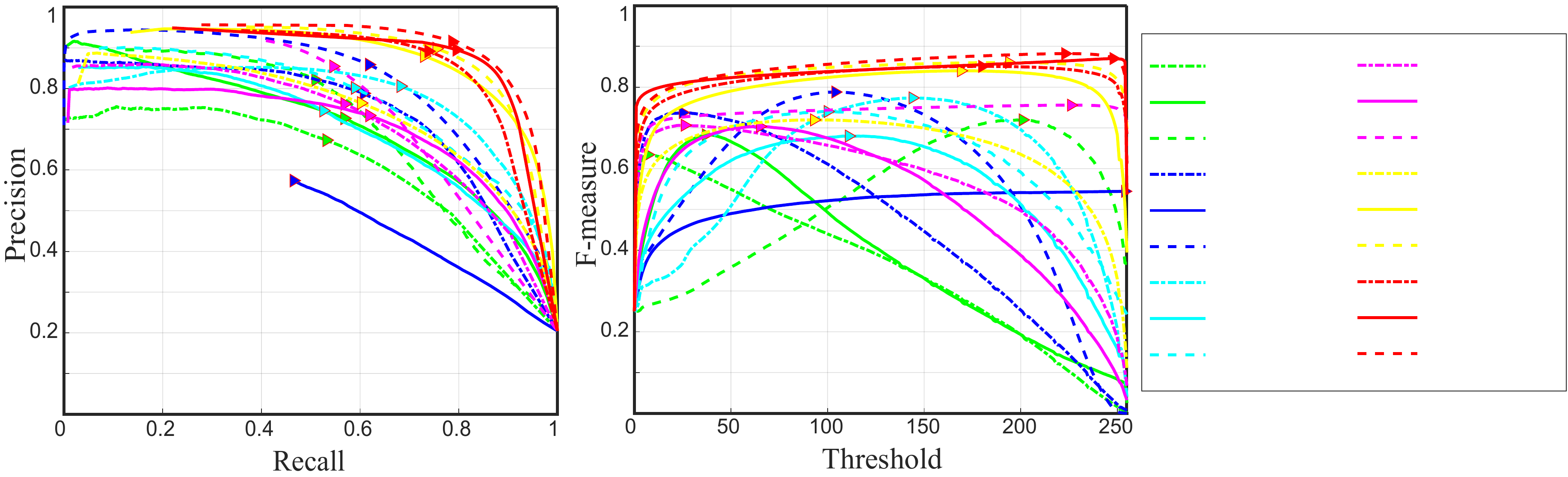}
    \put(90.6,25.50){SE \cite{guo2016salient}}
    \put(90.6,23.19){DF \cite{qu2017rgbd}}
    \put(90.6,20.88){AFNet \cite{wang2019adaptive}}
    \put(90.6,18.56){CTMF \cite{han2017cnns}}
    \put(90.6,16.25){MMCI \cite{chen2019multi}}
    \put(90.6,13.94){PCF \cite{chen2018progressively}}
    \put(90.6,11.63){TANet \cite{chen2019three}}
    \put(90.6,9.310){CPFP \cite{zhao2019Contrast}}
    \put(90.6,7.000){D3Net}
    \put(77.2,25.50){LHM \cite{peng2014rgbd}}
    \put(77.2,23.19){CDB \cite{liang2018stereoscopic}}
    \put(77.2,20.88){DESM \cite{cheng2014depth}}
    \put(77.2,18.56){GP \cite{ren2015exploiting}}
    \put(77.2,16.25){CDCP \cite{zhu2017innovative}}
    \put(77.2,13.94){ACSD \cite{ju2014depth}}
    \put(77.2,11.63){LBE \cite{feng2016local}}
    \put(77.2,9.310){DCMC \cite{cong2016saliency}}
    \put(77.2,7.000){MDSF \cite{song2017depth}}
  \end{overpic}
  \caption{\small PR Curve (Left) and F-measures (Right) under different thresholds on the proposed SIP dataset.
  }\label{fig:SIP_PRCurveAndFmeasure}
\end{figure*}

\subsection{Evaluation Metrics}\label{sec:metrics}
\textbf{MAE $M$.} We follow Perazzi \etal~\cite{Perazzi2012} \fdp{and}
evaluate the \textit{mean absolute error} (MAE) between a real-valued
saliency map $Sal$ and a binary ground truth $G$ for all image pixels:
\begin{equation}
    \begin{aligned}
    \text{MAE} = \frac{1}{N}|Sal-G|,
    \end{aligned}\label{eq:MAE}
\end{equation}
where $N$ is the total number of pixels.
The MAE estimates the approximation degree between the saliency map and the
ground truth map, and it is normalized to $[0, 1]$. The MAE provides a direct
estimate of conformity between estimated and ground truth maps.
However, for the MAE metric, small objects \fdp{are} naturally \fdp{assigned} smaller \fdp{errors, while}
larger objects \fdp{are given} larger errors. The metric \fdp{is} also \fdp{unable to} tell where the
error occurs~\cite{tsai2010motion}.

\textbf{PR Curve.} We also follow Borji~\etal~\cite{borji2015salient}
\fdp{and} provide the PR Curve. We divide \fdp{a} saliency map $S$ using a fixed
threshold which changes from 0 to 255.
For each threshold, a pair of recall \& precision scores are computed,
and \fdp{then} combined to form a precision-recall curve \fdp{that describes}
the model performance \fdp{in} different situations.
The overall evaluation results \fdp{for} PR Curves are shown in \figref{fig:PRCurveFmeasureOverall} (Top) and \figref{fig:SIP_PRCurveAndFmeasure} (Left).

\textbf{F-measure $F_{\beta}$.} F-measure is essentially a \fdp{region-based} similarity metric.
Following \fdp{the works by} Cheng and Zhang \etal~\cite{zhang2017amulet,borji2015salient},
we also provide the max F-measure using \fdp{various} fixed (0-255) thresholds.
The overall \fdp{F-measure} evaluation results under different thresholds on each
dataset are shown in \figref{fig:PRCurveFmeasureOverall} (Bottom) and \figref{fig:SIP_PRCurveAndFmeasure} (Right).

\begin{table*}[t!]
  \centering
  \renewcommand{\arraystretch}{1.10}
  \renewcommand{\tabcolsep}{0.34mm}
  \caption{\small
  Benchmarking results of 18 leading RGB-D approaches on our \emph{SIP} and fdp{six}
  classical~\cite{niu2012leveraging,li2014saliency,cheng2014depth,peng2014rgbd,ju2014depth,zhu2017three} datasets. 
  $\uparrow \& \downarrow$ denote larger and smaller is better, respectively.
  ``-T'' indicates the test set of the corresponding dataset.
  For traditional models, the statistics are based on overall datasets rather on the test set.
  The ``Rank'' denotes the ranking of each model in \fdp{a} specific measure.
  The ``All Rank'' indicates the overall ranking (average of each rank) in \fdp{a} specific dataset.
  The best performance is highlighted in \textbf{bold}.
  }\label{tab:BenchmarkResults}
  \small
  \begin{tabular}{lr||c|c|c|c|c|c|c|c|c|c|c||c|c|c|c|c|c|c}
  \hline\toprule
  &  &\multicolumn{11}{c||}{2014-2017}&\multicolumn{6}{c|}{2018-2019}&\multicolumn{1}{c}{} \\
  \cline{3-20}
   * & Model &
   LHM  & CDB  & DESM & GP    &
   CDCP & ACSD & LBE & DCMC & MDSF   & SE   & DF   & AFNet& CTMF & MMCI & PCF   & TANet& CPFP & \textbf{D$^3$Net}   \\
   &  & \cite{peng2014rgbd}        & \cite{liang2018stereoscopic}       & \cite{cheng2014depth}          & \cite{ren2015exploiting}              &
        \cite{zhu2017innovative}   & \cite{ju2014depth}                 & \cite{feng2016local}  & \cite{cong2016saliency}
        & \cite{song2017depth}   & \cite{guo2016salient} &\cite{qu2017rgbd}$^\dag$       & \cite{wang2019adaptive}$^\dag$ & \cite{han2017cnns}$^\dag$     & \cite{chen2019multi}$^\dag$
         & \cite{chen2018progressively}$^\dag$  &\cite{chen2019three}$^\dag$     & \cite{zhao2019Contrast}$^\dag$ & \textbf{Ours}$^\dag$   \\
  \midrule
   & Time (s) & 2.130 & - & 7.790 & 12.98 & $>$60.0 &0.718 & 3.110 & 1.200 & $>$60.0 & 1.570  &10.36  & 0.030& 0.630&0.050&0.060&0.070&0.170&\textbf{0.015}\\
   & Code  & M     & - & M  & M\&C    & M\&C &C   & M\&C& M     & C     &  M\&C&M\&C & Tf   & Caffe&Caffe&Caffe&Caffe&Caffe&Pytorch\\
  \midrule
  \midrule
  \multirow{4}{*}{\begin{sideways}\textit{NJU-T}\cite{ju2014depth}\end{sideways}}
    & $S_{\alpha}\uparrow$   & .514 &.624 & .665 & .527  & .669 & .699 & .695 & .686 & .748   & .664 & .763 & .772 & .849 & .858  & .877  & .878 & .879    & \trb{.900}      \\
    & $F_{\beta}\uparrow$     & .632 & .648 & .717 & .647  & .621 & .711 & .748& .715 & .775  & .748 & .804 & .775 & .845 & .852  & .872  & .874 & .877 & \trb{.900}      \\
    & $E_{\xi}\uparrow$       & .724 & .742 & .791 & .703  & .741 & .803 & .803& .799 & .838   & .813 & .864 & .853 & .913 & .915  & .924  & .925 & .926    & \trb{.950}      \\
    & $M\downarrow$ & .205 & .203 & .283 & .211  & .180 & .202 & .153& .172 & .157  & .169 & .141 & .100 & .085 & .079  & .059  & .060 & .053 & \trb{.041}      \\
    \toprule
    & $Rank$   & 17 & 16 & 14 & 17 & 15 & 12 & 10 & 13 & 9 & 11 & 7 & 7 & 6 & 5 & 4 & 3 & 2 & 1\\
    \midrule

  \multirow{4}{*}{\begin{sideways}\textit{STERE}\cite{niu2012leveraging}\end{sideways}}
    & $S_{\alpha}\uparrow$    & .562 & .615 & .642 & .588  & .713 & .692 & .660& .731 & .728   & .708 & .757 & .825 & .848 & .873  & .875   & .871 & .879 & \trb{.899}\\
    & $F_{\beta}\uparrow$     & .683 & .717 & .700 & .671  & .664 & .669 & .633 & .740& .719   & .755 & .757 & .823 & .831 & .863 & .860   & .861 & .874 & \trb{.891}\\
    & $E_{\xi}\uparrow$       & .771 & .823 & .811 & .743  & .786 & .806 & .787& .819 & .809  & .846 & .847 & .887 & .912 & .927& .925  & .923 & .925 & \trb{.938}\\
    & $M\downarrow$ & .172 & .166 & .295 & .182  & .149 & .200 & .250& .148 & .176  & .143 & .141 & .075 & .086 & .068  & .064   & .060 & .051 & \trb{.046}\\
    \toprule
    & $Rank$  & 16 & 12 & 14 & 18 & 13 & 15 & 17 & 10 & 11 & 9 & 8 & 7 & 6 & 3 & 4 & 5 & 2 & 1 \\

    \midrule
  \multirow{4}{*}{\begin{sideways}\textit{DES}\cite{cheng2014depth}\end{sideways}}
    & $S_{\alpha}\uparrow$    & .578 & .645 & .622 & .636  & .709 & .728 & .703 & .707& .741  & .741 & .752 & .770 & .863 & .848  & .842  & .858 & .872 & \trb{.898}\\
    & $F_{\beta}\uparrow$     & .511 & .723 & .765 & .597  & .631 & .756 & .788& .666 & .746  & .741 & .766 & .728 & .844 & .822  & .804   & .827 & .846 & \trb{.885}\\
    & $E_{\xi}\uparrow$       & .653 & .830 & .868 & .670  & .811 & .850 & .890& .773 & .851  & .856 & .870 & .881 & .932 & .928  & .893   & .910 & .923 & \trb{.946}      \\
    & $M\downarrow$ & .114 & .100 & .299 & .168  & .115 & .169 & .208& .111 & .122  & .090 & .093 & .068 & .055 & .065 & .049  & .046 & .038 & \trb{.031}\\
    \toprule
    & $Rank$   & 18 & 13 & 14 & 17 & 16 & 12 & 10 & 15 & 11 & 9 & 7 & 8 & 3 & 5 & 6 & 4 & 2 & 1\\

    \midrule
  \multirow{4}{*}{\begin{sideways}\textit{NLR-T}\cite{peng2014rgbd}\end{sideways}}
    & $S_{\alpha}\uparrow$    & .630 & .629 & .572 & .654  & .727 & .673 & .762& .724 & .805  & .756 & .802 & .799 & .860 & .856  & .874  & .886 & .888 & \trb{.912}\\
    & $F_{\beta}\uparrow$     & .622 & .618 & .640 & .611  & .645 & .607 & .745& .648 & .793 & .713 & .778 & .771 & .825 & .815  & .841  & .863 & .867 & \trb{.897}      \\
    & $E_{\xi}\uparrow$       & .766 & .791 & .805 & .723  & .820 & .780 & .855& .793 & .885   & .847 & .880 & .879 & .929 & .913  & .925  & .941 & .932 & \trb{.953}      \\
    & $M\downarrow$ & .108 & .114 & .312 & .146  & .112 & .179 & .081& .117 & .095 & .091 & .085 & .058 & .056 & .059  & .044  & .041 & .036 & \trb{.030}\\
    \toprule
    & $Rank$  & 14 & 15 & 16 & 18 & 12 & 17 & 10 & 13 & 7 & 11 & 8 & 8 & 5 & 6 & 4 & 3 & 2 & 1  \\
    \midrule

  \multirow{4}{*}{\begin{sideways}\textit{SSD}\cite{zhu2017three}\end{sideways}}
    & $S_{\alpha}\uparrow$    & .566 & .562 & .602 & .615  & .603 & .675 & .621& .704 & .673  & .675 & .747 & .714 & .776 & .813  & .841   & .839 & .807 & \trb{.857}\\
    & $F_{\beta}\uparrow$     & .568 & .592 & .680 & .740  & .535 & .682 & .619 & .711& .703   & .710 & .735 & .687 & .729 & .781  & .807   & .810 & .766 & \trb{.834}      \\
    & $E_{\xi}\uparrow$       & .717 & .698 & .769 & .782  & .700 & .785 & .736 & .786& .779  & .800 & .828 & .807 & .865 & .882  & .894   & .897 & .852 & \trb{.910}\\
    & $M\downarrow$ & .195 & .196 & .308 & .180  & .214 & .203 & .278& .169 & .192  & .165 & .142 & .118 & .099 & .082  & .062   & .063 & .082 & \trb{.058}      \\
    \toprule
    & $Rank$   & 16 & 17 & 15 & 11 & 17 & 13 & 14 & 9 & 12 & 9 & 7 & 8 & 6 & 4 & 2 & 2 & 5 & 1 \\

    \midrule
  \multirow{4}{*}{\begin{sideways}\textit{LFSD}\cite{li2014saliency}\end{sideways}}
    & $S_{\alpha}\uparrow$    & .553 & .515 & .716 & .635  & .712 & .727 & .729& .753 & .694   & .692 & .783 & .738 & .788 & .787  & .786 & .801 & \trb{.828} & .825      \\
    & $F_{\beta}\uparrow$     & .708 & .677 & .762 & .783 & .702 & .763 & .722 & .817 & .779   & .786 & .813 & .744 & .787 & .771  & .775   & .796 & \trb{.826} & .810      \\
    & $E_{\xi}\uparrow$       & .763 & .766 & .811 & .824  & .780 & .829 & .797& .856 & .819   & .832 & .857 & .815 & .857 & .839  & .827   & .847 & \trb{.872} & .862      \\
    & $M\downarrow$ & .218 & .225 & .253 & .190  & .172 & .195 & .214& .155 & .197   & .174 & .145 & .133 & .127 & .132  & .119   & .111 & \trb{.088} & .095\\
    \toprule
    & $Rank$ & 17 & 18 & 16 & 12 & 15 & 11 & 14 & 6 & 13 & 9 & 5 & 10 & 4 & 7 & 8 & 3 & 1 & 2  \\
    \midrule
   \multirow{4}{*}{\begin{sideways}\textit{SIP} (Ours)\end{sideways}}
    & $S_{\alpha}\uparrow$    & .511 & .557 & .616 & .588  & .595 & .732 & .727& .683 & .717   & .628 & .653 & .720 & .716 & .833  & .842   & .835 & .850 & \trb{.860}       \\
    & $F_{\beta}\uparrow$     & .574 & .620 & .669 & .687  & .505 & .763 & .751& .618 & .698   & .661 & .657 & .712 & .694 & .818  & .838   & .830 & .851 & \trb{.861}       \\
    & $E_{\xi}\uparrow$       & .716 & .737 & .770 & .768  & .721 & .838 & .853 & .743& .798   & .771 & .759 & .819 & .829 & .897 & .901   & .895 & .903 & \trb{.909}       \\
    & $M\downarrow$ & .184 & .192 & .298 & .173  & .224 & .172 & .200& .186 & .167   & .164 & .185 & .118 & .139 & .086  & .071   & .075 & .064 & \trb{.063}       \\
    \toprule
    & $Rank$  & 17 & 16 & 14 & 12 & 18 & 6 & 9 & 14 & 10 & 11 & 13 & 7 & 8 & 5 & 3 & 4 & 2 & 1  \\
    \toprule
    & $All~Rank$  & 18 & 17 & 15 & 14 & 16 & 13 & 12 & 11 & 10 & 9 & 7 & 8 & 6 & 5 & 4 & 3 & 2 & 1 \\
  \bottomrule
  \hline
  \end{tabular}
\end{table*}

\textbf{S-measure $S_{\alpha}$.} Both \fdp{the} MAE and F-measure metrics ignore important \fdp{structural}
information. \fdp{However}, behavioral vision studies have shown that the human visual system
is highly sensitive to structures in scenes~\cite{fan2017structure}.
Thus, we additionally include the structure measure (S-measure~\cite{fan2017structure}).
The S-measure combines the region-aware ($S_r$) and
object-aware ($S_o$) structural similarity as \fdp{the} final structure metric:
\begin{equation}
\label{equ:S-measure}
S_{\alpha} = \alpha*S_o+(1-\alpha)*S_r,
\end{equation}
where $\alpha\!\in\![0,1]$ is the balance parameter and set to 0.5.

\textbf{E-measure $E_{\xi}$.} E-measure is the recent\fdp{ly} proposed Enhanced alignment
measure~\cite{Fan2018Enhanced} \fdp{from} the binary map evaluation field.
This measure \fdp{is} based on cognitive vision studies, \fdp{and} combines local pixel values
with the image-level mean value in one term, jointly capturing image-level statistics
and local pixel matching information. Here, we introduce max/maximal E-measure to
provide a more comprehensive evaluation.

\subsection{Metric Statistics}\label{sec:metricStatistics}
For a given metric $\zeta \in \{S_\alpha,F_\beta,E_{\xi},M\}$ we consider different statistics.
%
$I_j^i$ denote \fdp{an} image \fdp{from a} specific dataset $D_i$.
Thus, $D_i = \{I_1^i,I_2^i,\ldots,I_j^i\}$.
Let $\overline{\zeta}(I^i_j)$ be the metric score on image $I^i_j$.
The \emph{mean} is the average dataset statistic defined
as $M_\zeta(D_i) = \frac{1}{|D_i|}\sum\overline{\zeta}(I^i_j)$, where
$|D_i|$ is the total number of images on the $D_i$ dataset.
The mean statistic\fdp{s} over different datasets are summarized
in \tabref{tab:BenchmarkResults}.

\begin{figure*}[t!]
  \centering
  \begin{overpic}[width=\textwidth]{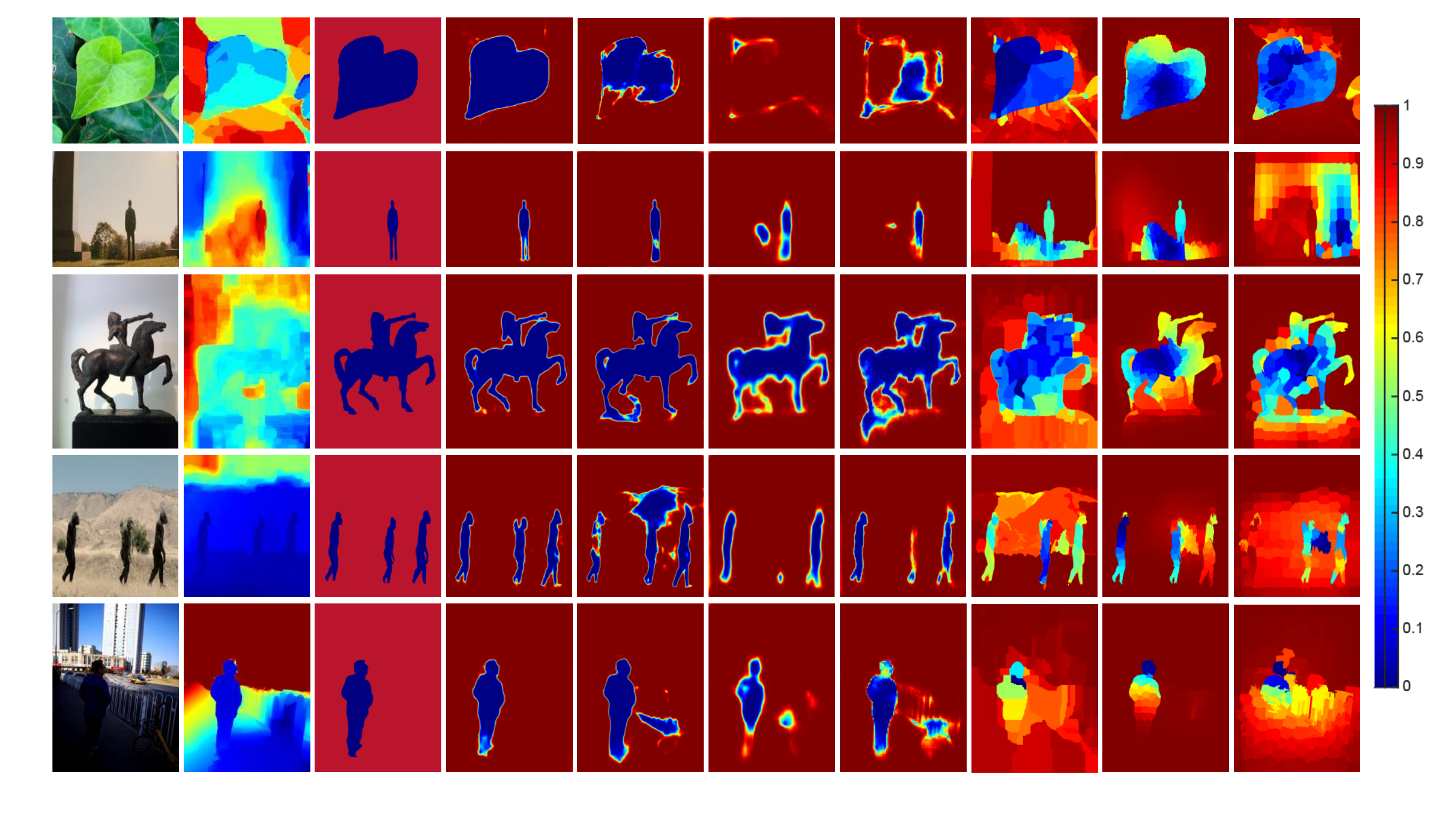}
  \put(5.5,2.5){\scriptsize (a) RGB}
  \put(14,2.5){\scriptsize  (b) Depth}
  \put(24,2.5){\scriptsize  (c) GT}
  \put(32,2.5){\scriptsize  (d) \textbf{D$^3$Net}}
  \put(39,2.5){\scriptsize  (e) CPFP\cite{zhao2019Contrast}}
  \put(48.5,2.5){\scriptsize  (f) TANet\cite{chen2019three}}
  \put(58.2,2.5){\scriptsize(g) PCF\cite{chen2018progressively}}
  \put(67,2.5){\scriptsize  (h) MDSF\cite{song2017depth}}
  \put(77,2.5){\scriptsize  (i) SE\cite{guo2016salient}}
  \put(85,2.5){\scriptsize  (j) DCMC\cite{cong2016saliency}}

  \put(1,49){\footnotesize \rotatebox{90}{\emph{LFSD}~\cite{li2014saliency}}}
  \put(1,39){\footnotesize \rotatebox{90}{\emph{NJU2K}~\cite{ju2014depth}}}
  \put(1,29){\footnotesize \rotatebox{90}{\emph{STERE}~\cite{niu2012leveraging}}}
  \put(1,18){\footnotesize \rotatebox{90}{\emph{SSD}~\cite{zhu2017three}}}
  \put(1,7){\footnotesize  \rotatebox{90}{\emph{SIP} (Ours)}}
  \end{overpic}
  \caption{\small
  Visual comparisons with \fdp{the top-3 CNN-based models} (CPFP~\cite{zhao2019Contrast},
  TANet~\cite{chen2019three}, and PCF~\cite{chen2018progressively}) and \fdp{three} classical non-deep \fdp{methods} (MDSF~\cite{song2017depth}, SE~\cite{song2017depth}
  and DCMC~\cite{cong2016saliency}), on five datasets.
  Further results \fdp{can be found} in \supp{\url{http://dpfan.net/D3NetBenchmark}}.
 }\label{fig:result}
\end{figure*}

\subsection{Performance Comparison and Analysis}\label{sec:PerformanceComparision}
\textbf{Performance of Traditional Models.}
Based on the overall performance\fdp{s} listed in \tabref{tab:BenchmarkResults},
we observe that ``\emph{SE~\cite{guo2016salient}, MDSF~\cite{song2017depth},
and DCMC~\cite{cong2016saliency} are \fdp{the top-3} traditional algorithms.}''
Utilizing superpixel technology, both SE and DCMC explicitly extract the
region contrast features from \fdp{an} RGB image. In contrast, MDSF formulate\fdp{s} SOD
as a pixel-wise binary labelling problem, which is solved by SVM.

\textbf{Performance of Deep Models.}
\fdp{Our} D$^3$Net, CPFP~\cite{zhao2019Contrast}
and TANet~\cite{chen2019three} are \fdp{the top-3}
deep models \fdp{out of all} leading methods,
showing the strong feature representation ability
of deep learning for this task.

\textbf{Traditional \emph{vs} Deep Models.}
From \tabref{tab:BenchmarkResults}, we observe that most of the deep models perform
better than the traditional algorithms. Interestingly, MDSF~\cite{zhu2017three}
outperforms two deep models (\ie, DF~\cite{qu2017rgbd} and AFNet~\cite{wang2019adaptive})
on \fdp{the} \emph{NLPR} dataset.


\subsection{Comparison with SOTAs}\label{sec:SOTAComparison}
We compare our D$^3$Net \fdp{with}
17 SOTA models in \tabref{tab:BenchmarkResults}.
In general, our model outperforms the best published result (CPFP~\cite{zhao2019Contrast}-CVPR'19)
by large margin\fdp{s} of 1.0\% $\sim$ 5.8\% on \fdp{six} datasets.
Notably, we \fdp{also} achieve \fdp{a} significant improvement of 1.4\% on
the proposed real-world \emph{SIP} dataset.

We also report saliency maps generated on various challenging
scenes to show the \fdp{visual} superiority of our D$^3$Net.
Some representative examples are shown in \figref{fig:result}, such as
\fdp{when the} structure of the salient object in \fdp{the} depth map
\fdp{is} partially (\eg, the $1^{st}$, $4^{th}$, and $5^{th}$ rows) or
dramatically (\ie, the $2^{nd}$-$3^{rd}$ rows) damaged.
\fdp{Specifically}, in the $3^{rd}$ and $5^{th}$ rows, the depth of the salient
object is locally connected with background scenes.
Also, the $4^{th}$ row contains multiple isolated salient objects.
For these challenging situations, most of \fdp{the} existing top competitors
are unlikely to locate the salient objects due to \fdp{their} poor
depth maps or insufficient multi-modal fusion schemes.
Although CPFP~\cite{zhao2019Contrast}, TANet~\cite{chen2019three},
and PCF~\cite{chen2018progressively} can generate more
correct saliency maps than others, the salient object often
introduces noticeable \fdp{distinct} backgrounds ($3^{rd}$-$5^{th}$ rows)
or \fdp{the} fine details of \fdp{the} salient object \fdp{are lost}($1^{st}$ row) due to
the \fdp{lack} of \fdp{a} cross-modality learning ability.
In contrast, our D$^3$Net can eliminate low-quality depth maps
and adaptively select complementary cues from RGB and \fdp{depth images}
to infer the real salient object and highlight its details.

\section{Applications}\label{sec:application}
\subsection{Human Activities}
\fdp{Nowadays, mobile phones generally have deep sensing cameras.
With RGB-D salient object detection, users can better achieve the following functions: object extraction, a bokeh effect, mobile user recognition, \etc.
Many monitoring probes also have depth sensors, and RGB-D SOD can be helpful to the discovery of suspicious objects.
For example, there is a lidar probe in autonomous vehicles designed to obtain depth information.
RGB-D SOD is thus helpful for detecting basic objects such as pedestrians and signboards in these vehicles.
There are also depth sensors in most industrial robots, so RGBD-SOD can help them better perceive the environment and take certain actions.
}

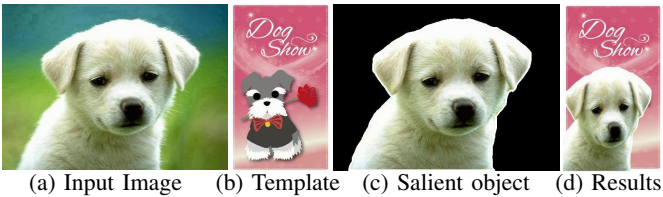
\begin{figure}[t!]
  \centering
  \small
  \begin{overpic}[width=\columnwidth]{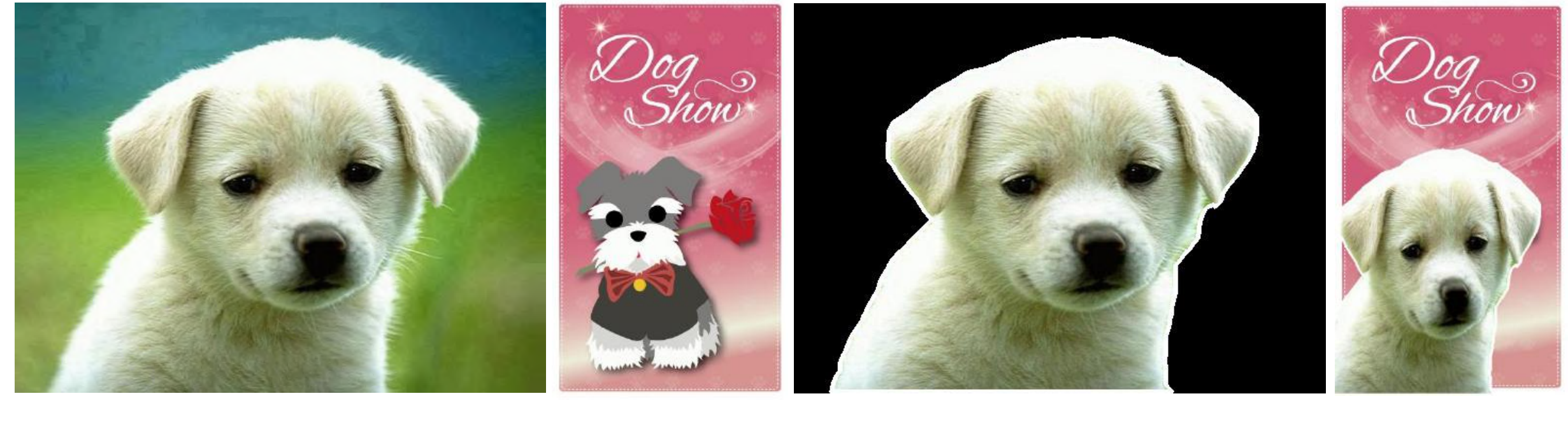}
  \put(5,0){(a) Input Image}
  \put(33,0){(b) Template}
  \put(55,0){(c) Salient object}
  \put(84,0){(d) Results}
  \end{overpic}
  \caption{\small
  Examples of book cover maker.
  See \secref{sec:application} for details.
 }\label{fig:BackgroundChange}
\end{figure}

\subsection{Background Changing Application}
Background \fdp{changing} techniques \fdp{have} become vital for art designers to
leverage the increasing volumes of available image database.
Traditional designers utilize photoshop to design
their products. This is quite \fdp{a} time-consuming task and
requires \fdp{significant} technical knowledge.
A large majority of potential users fail to grasp the \fdp{high-skilled} technique
in the art design. Thus, an easy-to-use application is needed.

To overcome the above-mentioned drawbacks, salient object detection
technology \fdp{could} be a potential solution.
Previous similar work\fdp{s}, such as \fdp{the} automatic generation of
visual-textual applications~\cite{yang2016automatic,jahanian2013recommendation}
motive us to create \fdp{a} background changing application for book cover layouts.
We \fdp{provide} a prototype demo, as shown in \figref{fig:BackgroundChange}.
First, the user can upload \fdp{an} image
\fdp{as} a candidate \fdp{design} image ((a) Input Image). Then, content-based image features, such as \fdp{an} RGB-D based saliency map, \fdp{are} considered in \fdp{order to automatically generate} salient object\fdp{s}. Finally, the system allows us to choose from
our library of professionally designed book cover layouts ((b) Template).
By combining high-level template constraints and low-level image features, we
obtain the background changed book cover ((d) Results). 

\begin{table*}[t!]
  \centering
  \small
  \renewcommand{\arraystretch}{1.0}
  \renewcommand{\tabcolsep}{2.6mm}
   \caption{\small
  S-measure$\uparrow$ score on our \emph{SIP} and \fdp{the} STERE dataset.
  The symbol $\uparrow$ indicates that the higher the score is, the better the model performs and vice versa.
  See details in \secref{sec:discussion}.
  }\label{tab:AblationStudy}
  \begin{tabular}{c|l||ccccccc}
  \hline\toprule
  Aspects & Model & \emph{SIP (Ours)} & \emph{STERE}~\cite{niu2012leveraging} & DES~\cite{cheng2014depth} & LFSD~\cite{li2014saliency} & SSD~\cite{zhu2017three} & NJU2K~\cite{ju2014depth}& NLPR~\cite{peng2014rgbd} \\
  \hline
  \hline
  \multirow{3}{*}{w/o DDU}
  & RgbNet      & 0.831 & 0.893  & 0.881 & 0.810 & 0.839 & 0.888 & 0.911\\
  & RgbdNet     & 0.862 & 0.898  & 0.896 & 0.836 & 0.857 & 0.898 & 0.910\\
  & DepthNet    & 0.862 & 0.713  & 0.911 & 0.724 & 0.811 & 0.857 & 0.864\\
  \hline
  \multirow{3}{*}{DDU}
  & Lower Bound              & 0.822 & 0.881 & 0.870 & 0.788 & 0.817 & 0.875 & 0.897\\
  & \textbf{D$^3$Net (Ours)} & \textit{0.860} & \textit{0.899} & \textit{0.898} & \textit{0.825} & \textit{0.857} & \textit{0.900} & \textit{0.912}\\
  & Upper Bound              & 0.872 & 0.910 & 0.907 & 0.858 & 0.879 & 0.912 & 0.924\\
  \hline\toprule
  \end{tabular}
\end{table*}

\begin{figure*}[t!]
  \centering
  \begin{overpic}[width=\textwidth]{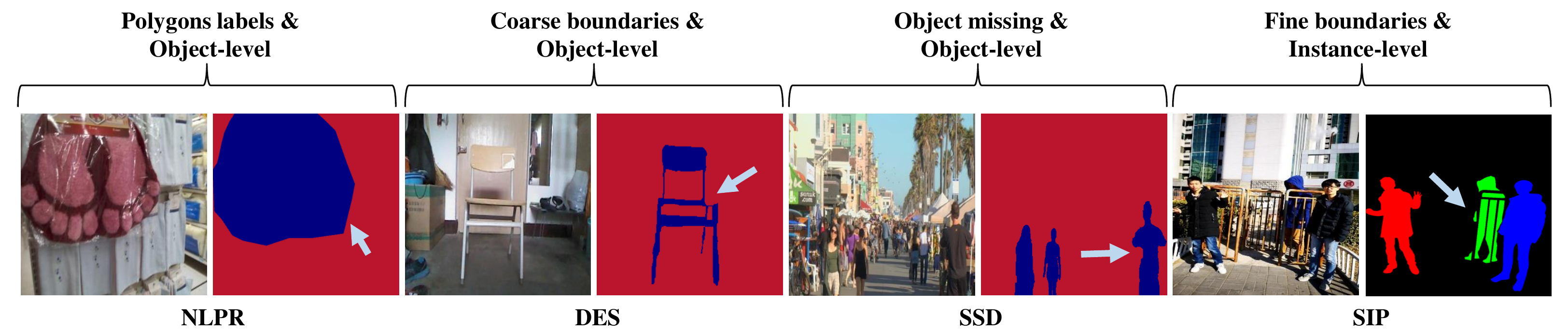}
  \end{overpic}
  \caption{\small Comparison with previous object-level datasets, which
  are labeled with polygons (the foot pad in \textit{NLPR}~\cite{peng2014rgbd}),
  coarse boundaries (\ie, the chair in \textit{DES}~\cite{cheng2014depth}), and
  missed object parts (\eg, the person in \textit{SSD}~\cite{zhu2017three}).
  \fdp{In contrast}, the proposed object-/instance-level~\emph{SIP}~dataset is labeled with smooth,
  fine boundaries.
  More specifically, occlusion\fdp{s are} also considered (\eg, the barrier region). 
  }\label{fig:labelQuality}
\end{figure*}

\fdp{Since designing a complete software system is not our main focus in this article,}
Future researchers can \fdp{follow} yang~\etal~\cite{yang2016automatic} \fdp{and} set our visual background
image with a specified topic~\cite{jahanian2013recommendation}.
In stage two, the input image \fdp{is} resized
to \fdp{match} the target style size and preserve the salient region according
to \fdp{the} inference of our D$^3$Net model.

\section{Discussion}\label{sec:discussion}

Based on our comprehensive benchmarking results, we present our
conclusions to the most important questions that may benefit the
research community to rethink the RGB-D image for salient object detection.

\subsection{Ablation Study.}\label{sec:ablationStudy}
\fdp{We now provide a detailed analysis on the proposed baseline D$^3$Net model.
To verify the effectiveness of the depth map filter mechanism (the DDU),
we derive two ablation studies: w/o DDU and DDU, which refer to our D$^3$Net without utilizing DDU or include the DDU. For w/o DDU, we further test the performance of the three sub-network in the test phase of D$^3$Net.
%
In \tabref{tab:AblationStudy}, we observe that RgbdNet performs better than RgbNet on the \emph{SIP},
\emph{STERE}, \emph{DES}, \emph{LFSD}, \emph{SSD}, \emph{NJU2K} datasets. It indicates that the
cross-modality (RGB and depth) features show strong promise for RGB-D image representation learning.
In most cases, however, DepthNet has lower performance than DepthNet and RgbNet.  It shows that only based on a single modality, it is difficult for the model to construct the structure of the geometry in an image.}
%
%

%
From \tabref{tab:AblationStudy}, we also observed that the use of the
DDU improves \fdp{the} performance (compared to RgbdNet) to a certain extent on the \emph{STERE}, \emph{DES}, \emph{NJU2K}, and \emph{NLPR} datasets.
We attribute the improvement to the DDU being able to discard low-quality depth maps and select one optimal path (RgbNet or RgbdNet).
For the SSD dataset, however, the DDU achieves comparable performance to the single stream network (\ie, RgbdNet).
It is worth mentioning that D$^3$Net outperforms any prior approach intended for SOD, without any post-processing techniques, such as CRF, which are typically used to boost scores. In order to know the lower and upper bound of our D$^3$Net, we additionally select the optimal path (RgbdNet or RgbNet) of the D$^3$Net. For example, for a specific RGB ($I_{rgb}$) and depth map ($I_{depth}$), the two predicted maps \ie, $S_{rgb}$ and $S_{rgbd}$, can be assessed separately. Thus, for each input we know the best output in existing network. We aggregate all the best and worst results and achieve the upper bound and lower bound of our D$^3$Net. From existing results listed in \tabref{tab:AblationStudy}, D$^3$Net still has a $\sim$1.6\% performance gap on average related to the upper bound.

\subsection{Limitations}\label{sec:limitation}
First, it is worth pointing out that the number of images in
the \emph{SIP} dataset is relatively \fdp{small} compared with most
datasets for RGB \fdp{salient object detection.}
Our goal \fdp{behind building} this dataset is to explore the potential direction
of \fdp{smartphone} based application\fdp{s}. As can be seen from the benchmark results and the
demo application described in \secref{sec:application}, salient object detection
over real human \fdp{activity} scenes is a promising direction.
We plan to keep growing the dataset with more challenging situations and various kinds of foreground person\fdp{s}.

Second, our simple general framework D$^3$Net consists of three  sub-networks, which may increase the memory on \fdp{a} light-weight device.
In \fdp{a} real environment, \fdp{several} strategies can be considered \fdp{to avoid this,}
such as \fdp{replacing} the backbone with MobileNet V2~\cite{sandler2018mobilenetv2}, \fdp{dimension reduction~\cite{zhang2018local}},
or \fdp{using the recently} released ESPNet V2~\cite{mehta2018espnetv2} models.
\fdp{Third, we present the lower and upper bounds of the DDU.
The optimal upper bound is obtained by feeding the input
into RgbdNet or RgbNet so that the predicted map is optimal.
As shown in \tabref{tab:AblationStudy}, our DDU module does not
achieve the best upper bound on the current training subset.
There \fdp{is thus} still an opportunity to design a better DDU to further improve the performance.}
%

\section{Conclusions}\label{sec:conclusion}
We present systematic studies on RGB-D based salient object detection
by: (1) Introducing a new human-oriented \emph{SIP} dataset
reflecting the realistic in-the-wild mobile use scenarios.
%
(2) Designing a novel D$^3$Net.
(3) Conducting so far the largest-scale ($\sim$97K) benchmark.
\fdp{Compared with existing datasets, \emph{SIP} covers several
challenges (\eg, background diversity, occlusion, \etc) of human in the real environments.}
Moreover, the proposed baseline achieves promising results.
It is among the fastest methods, making it a practical solution \fdp{to RGB-D salient object detection}.
%
The comprehensive benchmarking results include 32 summarized SOTAs and
18 evaluated traditional/deep models. We hope this benchmark \fdp{will} accelerate not only the development of this area
but also others (\eg, stereo estimating/matching~\cite{nie2019Multi}, multiple salient person
detection, salient instance detection~\cite{li2017instance}, \fdp{sensitive object detection~\cite{yu2016iprivacy}}, \fdp{image segmentation~\cite{shen2019submodular}}).
\fdp{Note that the methods utilized in our D$^3$Net baseline are simple and more complex components (\eg, PDC in~\cite{fan2020Camouflage}) or training strategy~\cite{fan2020inf} are promising to increase the performance.
In the future, we plan to incorporate recently proposed techniques \eg, the weighted triplet loss~\cite{yu2019spatial}, hierarchical deep features~\cite{yu2019hierarchical}, visual question-driven saliency~\cite{he2019exploring}, into our D$^3$Net to further boost the performance.}
After this submission, there are many interesting models, such as UCNet~\cite{zhang2020UCNet}, JL-DCF~\cite{Fu2020JLDCF}, GFNet~\cite{liu2020cross},
DMRA~\cite{piao2019depth}, ERNet~\cite{piaoexploit}, BiANet~\cite{zhang2020bilateral}, \etc, have been released. Please refer to our online leaderboard (\href{http://dpfan.net/d3netbenchmark/}{http://dpfan.net/d3netbenchmark/}) for more details. This website will be updated continually.
We foresee this study \fdp{driving} salient object detection
towards real-world application scenarios with 
multiple salient persons and complex interactions
through the mobile device (\eg, \fdp{smartphone or tablets}).

\vspace{.15in}
\noindent\textbf{Acknowledgment. \quad}
We thank Jia-Xing Zhao, Yun Liu, and Qibin Hou for insightful feedback.
This research was supported by Major Project for New Generation of AI
under Grant No. 2018AAA0100400, NSFC (61922046),
and Tianjin Natural Science Foundation (17JCJQJC43700).

\ifCLASSOPTIONcaptionsoff
  \newpage
\fi

{\small
\bibliographystyle{IEEEtran}
\bibliography{rgbdDataset}
}

\vspace{-.3in}
\begin{IEEEbiography}[{\includegraphics[width=1in,keepaspectratio]{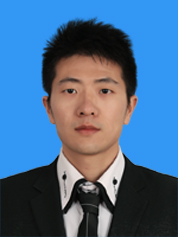}}]
{Deng-Ping Fan} received his PhD degree from Nankai University of Tianjin in 2019.
He joined Inception Institute of Artificial Intelligence (IIAI), UAE in 2019.
From 2015 to 2019, he was a Ph.D. candidate in Department of Computer Science, University
of Nankai, directed by Prof. Ming-Ming Cheng. He received the Huawei Scholarship in 2017.
His current research interests include computer vision, image processing and deep learning.
\end{IEEEbiography}
\vspace{-.3in}

\begin{IEEEbiography}[{\includegraphics[width=1in,keepaspectratio]{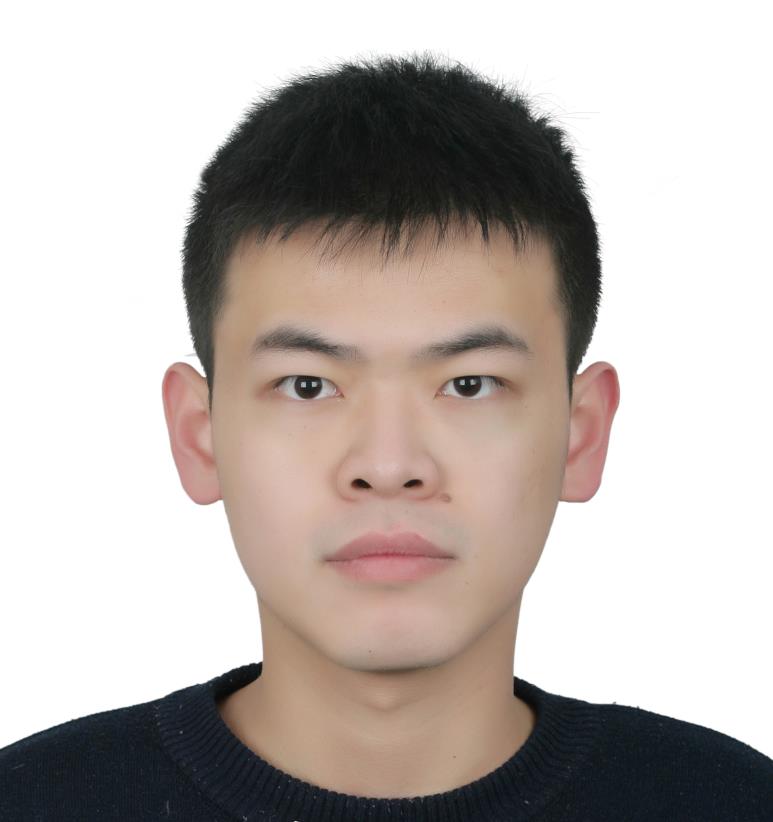}}]
{Zheng Lin} is currently a Ph.D. candidate with
College of Computer Science, Nankai University, under the supervision of
Prof. Ming-Ming Cheng. His research interests include deep learning,
computer graphics and computer vision.
\end{IEEEbiography}
\vspace{-.3in}

\begin{IEEEbiography}[{\includegraphics[width=1in,keepaspectratio]{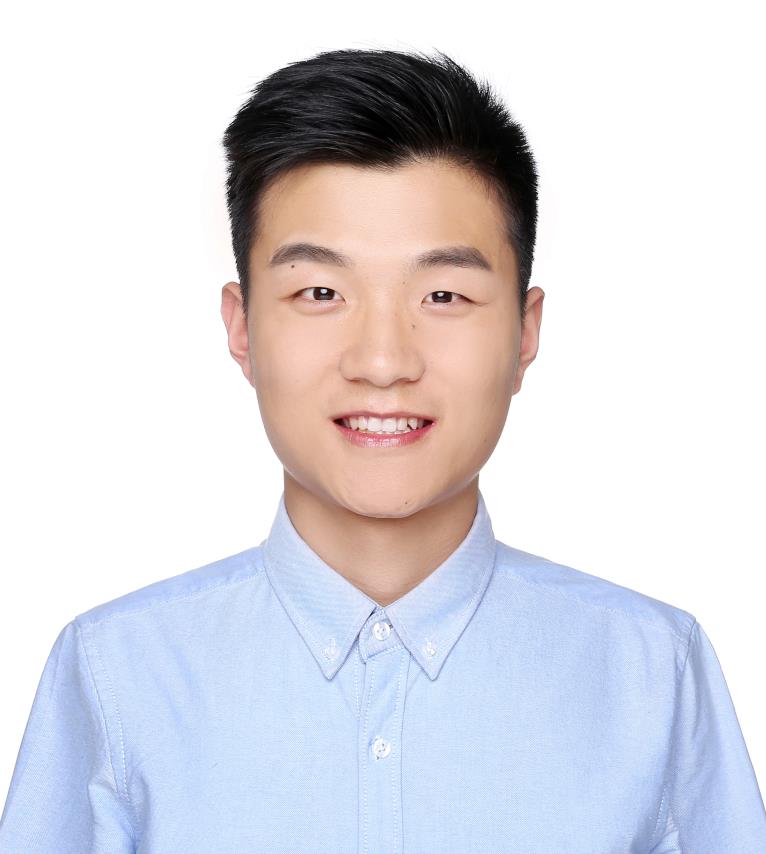}}]
{Zhao Zhang} received the B.Eng degree from Yangzhou University in 2019.
Currently, he is a master student in Nankai University under the
supervision of Prof. Ming-Ming Cheng. His research interests
includes computer vision and image processing.
\end{IEEEbiography}
\vspace{-.3in}

\begin{IEEEbiography}[{\includegraphics[width=1in,keepaspectratio]{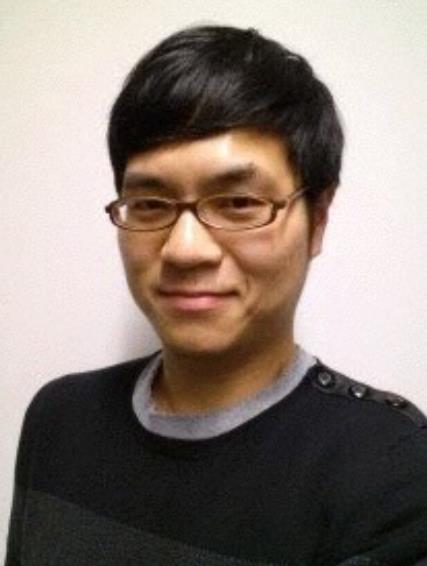}}]
{Menglong Zhu} is a Computer Vision Software
Engineer at Google. He obtained a Bachelor’s
degree in Computer Science from Fudan University, in 2010, and a Master’s degree in Robotics
and a PhD degree in Computer and Information Science from University of Pennsylvania, in
2012 and 2016, respectively. His research interests are on object recognition, 3D object/human
pose estimation, human action recognition, visual SLAM and text recognition.
\end{IEEEbiography}
\vspace{-.3in}

\begin{IEEEbiography}[{\includegraphics[width=1in,keepaspectratio]{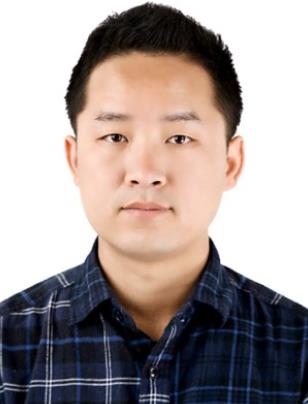}}]
{Ming-Ming Cheng } received his PhD degree
from Tsinghua University in 2012. Then he did 2
years research fellow, with Prof. Philip Torr in Oxford.
He is now a professor at Nankai University,
leading the Media Computing Lab. His research
interests includes computer graphics, computer
vision, and image processing. He received research
awards including ACM China Rising Star
Award, IBM Global SUR Award, CCF-Intel Young
Faculty Researcher Program, \etc.
\end{IEEEbiography}


\vfill


\end{document}